\documentclass[acmsmall]{acmart}

\acmJournal{TIST}

\usepackage{graphicx}
\usepackage{multirow}
\usepackage{algorithm}
\usepackage{algpseudocode}
\usepackage{amsmath}
\usepackage{graphicx}
\usepackage{subcaption}
\usepackage{booktabs}
\usepackage{tabularx}
\usepackage{array}
\usepackage{makecell}

\usepackage[table]{xcolor}
\usepackage{pgf}

\usepackage[edges]{forest}
\usetikzlibrary{
  arrows.meta,
  calc,
  decorations.pathreplacing,
  shapes.geometric,
  shapes.misc
}
\usepackage{fontawesome5}

\definecolor{negbase}{RGB}{148,50,50}
\definecolor{posbase}{RGB}{55,155,95}

\newcommand{\negd}[1]{%
  \pgfmathtruncatemacro{\pct}{
    min(30,max(2,abs(#1)/40*30))
  }%
  \edef\applyshade{%
    \noexpand\cellcolor{negbase!\pct!white}
  }%
  \applyshade #1%
}

\newcommand{\posd}[1]{%
  \pgfmathtruncatemacro{\pct}{
    min(30,max(2,abs(#1)/5*30))
  }%
  \edef\applyshade{%
    \noexpand\cellcolor{posbase!\pct!white}
  }%
  \applyshade +#1%
}

\title{Early Stopping in Large Reasoning Models:
A Survey and Empirical Audit of Efficiency
and Reliability}

\author{Quang Hung Nguyen}
\affiliation{%
  \institution{VinUniversity}
  \city{Hanoi}
  \country{Vietnam}
}
\email{hung.nq3@vinuni.edu.vn}

\author{Hanh Nguyen}
\affiliation{%
  \institution{VinUniversity}
  \city{Hanoi}
  \country{Vietnam}
}
\email{hanh.nth@vinuni.edu.vn}

\author{Thanh Tam Nguyen}
\affiliation{%
  \institution{Griffith University}
  \city{Gold Coast}
  \country{Australia}
}
\email{t.nguyen19@griffith.edu.au}

\author{Quoc Viet Hung Nguyen}
\affiliation{%
  \institution{Griffith University}
  \city{Gold Coast}
  \country{Australia}
}
\email{henry.nguyen@griffith.edu.au}

\author{Trinh Pham}
\affiliation{%
  \institution{VinUni Bigdata Research Institute,
  VinUniversity}
  \city{Hanoi}
  \country{Vietnam}
}
\email{phkhanhtrinh23@gmail.com}

\author{Thanh Trung Huynh}
\affiliation{%
  \institution{VinUniversity}
  \city{Hanoi}
  \country{Vietnam}
}
\email{trung.ht@vinuni.edu.vn}

\begin{document}

% ...

\begin{abstract}

Large reasoning models improve problem-solving accuracy through extended reasoning, but the resulting token cost motivates methods that stop generation before the reasoning trace is complete. However, reducing reasoning length does not necessarily yield efficient and reliable answers: stopping decisions introduce additional computation, and truncated reasoning may not adequately support the final answer. In this paper, we review early-stopping methods according to the information used to determine when to stop and conduct an empirical audit of ten methods across five reasoning models and eight benchmarks. We use a common replay protocol to compare methods on the same reasoning trajectories and account for tokens generated during both verification and answering. We introduce Target-Integrated Token Efficiency (TITE) to summarize accuracy gains and token cost relative to direct answering and a discrete-budget hindsight oracle. Our results show that the evaluated methods struggle to balance accuracy and token cost, particularly on coding and factuality tasks, with no method consistently dominating across settings. Randomized exit experiments on selected models and benchmarks reveal correctness reversals exceeding those predicted by a fitted decoding-noise baseline. Reliability analyses further identify signs of continued reasoning after forced termination and weaker justification in some correctly answered cases. These findings highlight the need to assess early stopping jointly through token cost, correctness stability, and the adequacy of retained reasoning.

\end{abstract}

\maketitle



\section{Introduction}

Large reasoning models (LRMs) have achieved substantial improvements in complex problem-solving by allocating additional computation at inference time. Rather than producing an answer directly, these models generate intermediate reasoning steps to explore possible solutions, verify conclusions, and correct mistakes \cite{wei2022chain,wang2023selfconsistency}. Recent models, including OpenAI o1 \cite{openai2026openaio1card} and DeepSeek-R1 \cite{deepseekai2025deepseekr1incentivizingreasoningcapability}, have demonstrated the effectiveness of this approach. However, stronger reasoning comes at a considerable computational cost. LRMs may generate lengthy reasoning traces, revisit conclusions, or continue reasoning even after obtaining sufficient information to answer correctly \cite{chen2025think23overthinkingo1like,sui2025stopoverthinkingsurveyefficient}. This creates a fundamental challenge: how can we reduce unnecessary reasoning without compromising the quality of the final answer?

Early-stopping methods address this challenge by identifying when the reasoning generated so far is sufficient to produce a correct answer. Existing approaches use signals such as answer convergence, confidence, entropy, and semantic redundancy to determine when to terminate reasoning \citep{yang2026dynamic,liu2025answerconvergencesignalearly,wang2026entropythinkreasoningmodel}. However, three aspects of early stopping remain insufficiently understood. First, existing methods may overlook verification costs and fail to generalize across models and tasks. Second, continued reasoning can turn a correct answer into an incorrect one, challenging the monotonicity assumption that exists in several early-stopping methods \citep{han-etal-2025-token,wang2025efficientlongcotreasoning,wu-etal-2025-thought,ballon2026probingtrajectoriesreasoningtraces,song2026thinkbrakeefficientreasoninglogprobability,caldarella2026thinkingpastanswerevaluating,datta-etal-2026-llms}. Third, stopping too early may interrupt reasoning before sufficient justification for the final answer has been established. Together, these issues raise a fundamental question: how much reasoning computation can existing early-stopping methods reliably save without compromising answer quality?

These limitations highlight the need for a systematic evaluation that goes beyond conventional accuracy and token-reduction measurements. In this work, we investigate early-stopping methods along three complementary dimensions: efficiency, stability, and reliability. We assess how closely existing methods approach hindsight-oracle reasoning costs after accounting for stopping overhead, examine whether correctness remains stable across reasoning positions, and evaluate whether forced exits terminate reasoning cleanly while preserving adequate justification. We further study how these behaviors vary across model families and task domains. Through this evaluation, we aim to identify both the practical benefits and the fundamental limitations of existing early-stopping strategies.

Specifically, we address the following research questions:

\begin{itemize}
    \item \textbf{RQ1 (Efficiency):} How much computation do existing early-stopping methods save across models and tasks? (\ref{subsec:rq1_efficiency})

    \item \textbf{RQ2 (Stability):} How does answer correctness change as reasoning progresses? (\ref{subsec:rq2_preservation})

    \item \textbf{RQ3 (Reliability):} Do forced early exits terminate reasoning cleanly and preserve sufficient justification for the final answer? (\ref{subsec:rq4_reliability})
\end{itemize}

Our main contributions are summarized as follows:

\begin{itemize}
    \item \textbf{Efficiency evaluation.} We introduce TITE, an oracle-relative efficiency metric that accounts for the computational overhead of early-stopping decisions, enabling a more comprehensive comparison of token savings.

    \item \textbf{Reasoning stability analysis.} We investigate correctness across randomized exit positions to characterize non-monotonic reasoning behavior and distinguish systematic correctness changes from stochastic decoding variation.

    \item \textbf{Cross-model and cross-domain evaluation.} We systematically examine the effectiveness and generalization of existing early-stopping methods across multiple model families and task domains.

    \item \textbf{Forced-exit reliability analysis.} We extend evaluation beyond final-answer correctness by measuring reasoning continuation after forced exits and the adequacy of retained reasoning justifications.
\end{itemize}


\section{Early Exit Framework}

\subsection{Scope and Definition}

Early-stopping is a method that reduces inference cost by terminating an ongoing reasoning trace before its natural completion. Given a prompt $x$, a reasoning model $M$ generates a reasoning trace $r_{T}$ and a final answer $y^{\mathrm{full}}$:
$$
(r_{T},y^{\mathrm{full}})\sim M(x),
$$

where $T$ is the natural end of the reasoning process. An early-stopping method selects a stopping position $t<T$, appends end-of-thinking intervention $\tau$, and produces an answer from that prefix:
$$
y^{(t)}
\sim
M\left(
\cdot
\mid
x,r_{t}\circ\tau
\right).
$$

A method is considered \emph{prefix-preserving} when the reasoning tokens generated before position $t$ are identical to the corresponding prefix of the natural reasoning trajectory. We limit our study to inference-time early stopping within a single reasoning model $M$.

\subsection{Generic Early-stopping Mechanism}

A generic early-stopping method $m$ is represented by
\vspace{-2pt}
$$
\left(
S_m,\pi_m,G_m,\tau_m
\right)
$$
where $S_m$ is a checkpoint function, $\pi_m$ is a verifier that estimates whether the current prefix is sufficient for final answering, $G_m$ is a stopping rule from verifier and $\tau_m$ is the stopping intervention, such as an end-of-thinking token.

At reasoning position $t$, the checkpoint function observes the current prefix $r_{t}$:
$$
s_{m,t}=
S_m
\left(
x,
r_{t}
\right),
\qquad
s_{m,t}\in\{0,1\}.
$$

If $s_{m,t}=0$, the model continues reasoning. If $s_{m,t}=1$, the verifier evaluates the current state:
$$
u_{m,t} =
\pi_m
\left(
x,r_{t}
\right).
$$

The verifier output $u_{m,t}$ may be a confidence score, entropy value, tentative answer, agreement score, redundancy estimate, or another measure of reasoning sufficiency. Some verifiers can be computed directly from the reasoning trace, while others require temporary probe generations. The stopping decision is decided by
$$
d_{m,t} =
G_m
\left(
u_{m,t},r_{t}
\right),
\qquad
d_{m,t}\in\{0,1\}.
$$

If $d_{m,t}=0$, the reasoning continues. Else, the method applies $\tau_m$ and generates the final answer:
$$
y_m
\sim
M
\left(
\cdot
\mid
x,r_{t}\circ\tau_m
\right).
$$

Algorithm~\ref{alg} summarizes this process.

\begin{algorithm}[t]
\small
\caption{Generic Early-stopping Mechanism}
\label{alg}
\begin{algorithmic}[1]

\State $r \gets \emptyset$ \Comment{Initialize reasoning prefix}

\For{$t = 1$ to $T$}
    \State $z_t \sim M(\cdot \mid x,r)$
        \Comment{Generate next token}
    \State $r \gets r \circ z_t$
        \Comment{Update reasoning prefix}

    \If{$S_m(x,r) = 1$}
        \State $u_{m,t} \gets \pi_m(x,r)$
            \Comment{Compute verifier output}

        \If{$G_m(u_{m,t},r) = 1$}
            \State $r \gets r \circ \tau_m$
                \Comment{Apply stopping intervention}
            \State $y_m \sim M(\cdot \mid x,r)$
                \Comment{Generate final answer}
            \State \Return $(r,y_m,t)$
        \EndIf
    \EndIf
\EndFor

\State $y_m \sim M(\cdot \mid x,r)$
    \Comment{Answer after full reasoning}
\State \Return $(r,y_m,T)$

\end{algorithmic}
\end{algorithm}

\subsection{Conditions for a Clean Early-stopping Solution}

We define a clean solution through three conditions:

\paragraph{Correctness.}

The final answer produced from the selected reasoning prefix should be correct. With reference answer $g$, let $V$ be the scoring function. The correctness condition is
$
V\left(y_m,g\right)=1.
$

\paragraph{Efficiency.}

The method should reduce complete inference cost relative to natural full reasoning. Let $C_m$ denote the total cost of method $m$ and $C_{\mathrm{full}} $ denote the cost of natural full reasoning and its final answer. The efficiency condition is
$
C_m<C_{\mathrm{full}}.
$

\paragraph{Clean termination.}

After the stopping intervention is applied, the model should transition from reasoning to final-answer production without continuing unresolved derivation, search, verification, or self-correction in the answer channel.

\subsection{Assumption}

\paragraph{Assumption of ideal KV cache implementation}
Output-based verifiers (Section~\ref{subsec:taxonomy}) often add scoring or generating prompt at the end of the existing reasoning text, which can be optimized with previous KV Cache generation. We do not account for the computational overhead associated with reading, loading, storing, or managing the KV cache, as all results assume an ideal KV-cache management implementation. We report the total number of generated tokens, including any additional tokens required to evaluate the stopping condition, which can be used for approximating the actual consumption of prefill and decode state.

\paragraph{Assumption of runtime optimization}
Wall-clock runtime depends on model architecture, hardware characteristics, kernel optimizations, and implementation choices such as batching and scheduling. We therefore report generated-token counts as a more implementation-independent measure of reasoning efficiency.

\subsection{Failure Modes}

\paragraph{Stopping before answer readiness.}
The verifier may accept a reasoning prefix before the
model has derived sufficient information to produce a
reliable answer. Forcing the transition to final
answering at this point can prevent useful computation
or later error correction, turning a potentially
correct full-reasoning answer into an incorrect
early-stopping answer. This failure corresponds to the \emph{exit-early}
trap examined in Section~\ref{subsec:rq2_preservation}.

\paragraph{Post-stop reasoning leakage.}

Reasoning models are designed to finish a reasoning trace before generating the final answer. Appending an end-of-thinking token at an unusual position changes the model's generation context and may fail to terminate reasoning cleanly. The model may continue deriving equations, testing alternatives, verifying intermediate steps, or revising its conclusion in the final-answer channel \cite{wang2026pathlockexpertseparatingreasoning}. This behavior violates the clean-termination condition.

\subsection{Taxonomy}
\label{subsec:taxonomy}

Early-stopping methods differ mainly in the information used to determine
whether the current reasoning prefix is sufficient for answering. We therefore
classify methods according to the evidence consumed by the stopping verifier
$\pi_m$, while separately noting the probe mechanism, stopping policy, training requirement, and whether the method preserves the original reasoning trajectory.

\paragraph{Internal-state based verifiers.}
These methods estimate stopping readiness from representations already
available during model inference, most commonly hidden states associated with
the current reasoning prefix. A lightweight classifier can be trained to
predict whether the current state contains sufficient information to answer,
as in SpecExit, Thought Calibration, OS-Pruner, and Lynx
\citep{yang2025specexitacceleratinglargereasoning,
wu-etal-2025-thought,
ehab2026osprunerpruningchainsofthoughtreasoning,
akgül2025lynxlearningdynamicexits}.
TERMINATOR instead introduces a larger learned prediction module
\citep{nagle2026terminator}. Other works focus on other internal representation of LLM, with NEAT \citep{liu-etal-2026-neat} identifies neurons associated with natural reasoning termination and monitors them, while ASAG \citep{li2026stopreasoningwonthelp} derives stopping signals from attention states.

These approaches can make stopping decisions
without generating a provisional answer, making their inference-time
verification relatively inexpensive. Their stopping signal, however, depends
on how well the learned representation transfers across models, tasks, and
reasoning distributions.

\paragraph{Output-based verifiers.}
Within output-based verifiers, we further distinguish
answer-based and answer-free methods according to
whether verification requires generating a complete
provisional task answer.

\textbf{Answer-based verifiers.}
Answer-based methods explicitly generate provisional answers from intermediate
reasoning prefixes and use agreement or convergence among these answers as
evidencearly-stopping. Answer Convergence measures the
stabilization of intermediate answers
\citep{liu2025answerconvergencesignalearly}. DynaSor \cite{fu2025efficiently} similarly uses consistency
among intermediate outcomes, while PUMA \citep{min2026stopreasoningconvergessemanticpreserving} combines provisional-answer
information with information from the preceding reasoning trajectory.
Because these methods inspect the objective outputs, their stopping signal has a direct interpretation; however, they introduce substantial
additional token costs, where a valid provisional answer may contain thousands of tokens or unable to be directly compared.

\textbf{Answer-free verifiers.}
Answer-free methods attempt to estimate reasoning sufficiency without producing
a complete provisional answer. One family uses statistics available from the
current generation, such as token probabilities, entropy, uncertainty, or
transition-related signals. Statistical Early Stopping uses uncertainty-based
statistics \citep{xie2025statistical}; DEER \cite{yang2026dynamic} with its variance CoDE-Stop \cite{hosseini2026earlystoppinglargereasoning} and SABER \cite{cheng2026saberstabilityawareearlyexit} evaluate short
continuations around reasoning-transition points with token probabilities; and EAT \cite{wang2026entropythinkreasoningmodel}, ThinkBrake \cite{song2026thinkbrakeefficientreasoninglogprobability} and
EntroCut \cite{yan2026entrocutentropyguidedadaptivetruncation} use entropy-related signals.
A second family directly queries whether the current prefix is sufficient.
DTSR \cite{xiang-etal-2026-thinking} and FlashThink \cite{jiang2025flashthinkearlyexitmethod}, for example, use short sufficiency or verification probes
rather than requiring a complete task answer. These approaches usually require
more computation than a verifier operating only on existing internal states, but
substantially less than repeatedly generating full provisional answers.

\paragraph{Reasoning-path optimization.}
Unlike prefix-preserving early stopping, reasoning-path
optimization modifies the generation trajectory to reduce
unnecessary computation. NoWait suppresses reflection
tokens \citep{wang-etal-2025-wait}, while TokenSkip
compresses CoT by skipping less important tokens
\citep{xia-etal-2025-tokenskip}. NCoTS searches for
shorter reasoning paths
\citep{ling-etal-2026-neural}, and TokenProbe uses
token-level confidence signals to preserve informative
reasoning during reinforcement learning
\citep{zeng2026token}. Other methods intervene directly
in hidden representations: Activation-Steered Compression
shifts generation toward concise reasoning
\citep{azizi2025activationsteeringchainofthoughtcompression},
Latent Reward Steering adaptively corrects fragile states
\citep{li2026latent}, and constrained latent-space
optimization redirects generation toward efficient
reasoning trajectories \citep{hou2026efficient}.
Answer Convergence also modifies the probability of
emitting the end-of-thinking token to encourage earlier
termination
\citep{liu2025answerconvergencesignalearly}.
NEAT \citep{liu-etal-2026-neat} is a boundary case because strong activation evidence triggers a genuine
early exit, whereas weaker evidence suppresses reflection tokens and therefore
changes the continuation trajectory.
Because these interventions alter the generation process
rather than simply selecting a stopping point along an
unchanged trajectory, we discuss them as related
efficiency approaches but exclude them from our
prefix-preserving early-stopping comparison.

\begin{table}[t]
\centering
\footnotesize
\caption{\textbf{Comparison with closely related work.}
Checkmarks denote evaluation components explicitly supported by the study.
$\triangle$ denotes related evidence that does not meet the exact criterion.}
\label{tab:closest_work}

\setlength{\tabcolsep}{2.5pt}
\renewcommand{\arraystretch}{0.95}
\begin{tabular}{@{}l*{5}{c}*{4}{c}@{}}
\toprule
& \multicolumn{5}{c}{\textbf{Task Coverage}}
& \multicolumn{4}{c}{\textbf{Evaluation Dimensions}} \\
\cmidrule(lr){2-6}\cmidrule(lr){7-10}
\textbf{Study}
& Math & \makecell{Sci. QA} & \makecell{Gen. QA}
& \makecell{Complex\\Code}  & Fact.
& Oracle & \makecell{Exit\\Dynamics}
& \makecell{Post-stop\\Audit} & \makecell{CoT\\Quality} \\
\midrule
TALE~\cite{han-etal-2025-token}
& $\checkmark$ & & & & & $\triangle$ & $\triangle$ & & \\

Efficient Long CoT~\cite{wang2025efficientlongcotreasoning}
& $\checkmark$ & & & & & $\triangle$ & & & $\checkmark$ \\

Thought Calibration~\cite{wu-etal-2025-thought}
& $\checkmark$ & $\checkmark$ & & & & & & & \\

S-GRPO~\cite{dai2025sgrpoearlyexitreinforcement}
& $\checkmark$ & $\checkmark$ & & & & & $\triangle$ & & \\

\midrule

Answer Convergence~\cite{liu2025answerconvergencesignalearly}
& $\checkmark$ & $\checkmark$ & $\checkmark$ & & & & $\checkmark$ & & $\checkmark$ \\

DynaSor~\cite{fu2025efficiently}
& $\checkmark$ & & & & & $\triangle$ & $\checkmark$ & & \\

DEER~\cite{yang2026dynamic}
& $\checkmark$ & $\checkmark$ & & $\checkmark$ & & & & & \\

ThinkBrake~\cite{song2026thinkbrakeefficientreasoninglogprobability}
& $\checkmark$ & $\checkmark$ & & & & $\checkmark$ & $\checkmark$ & & \\

Probing Trajectories~\cite{ballon2026probingtrajectoriesreasoningtraces}
& & $\checkmark$ & $\checkmark$ & & & & $\checkmark$ & & \\

PUMA~\cite{min2026stopreasoningconvergessemanticpreserving}
& $\checkmark$ & $\checkmark$ & & $\checkmark$ & & & $\checkmark$ & & $\checkmark$ \\

\midrule

Thinking Past the Answer~\cite{caldarella2026thinkingpastanswerevaluating}
& $\checkmark$ & $\checkmark$ & & & & $\checkmark$ & $\checkmark$ & & \\

DTSR~\cite{xiang-etal-2026-thinking}
& $\checkmark$ & $\checkmark$ & & & & & & & \\

CoDE-Stop~\cite{hosseini2026earlystoppinglargereasoning}
& $\checkmark$ & $\checkmark$ & & & & $\triangle$ & & & \\

EntroCut~\cite{yan2026entrocutentropyguidedadaptivetruncation}
& $\checkmark$ & & & & & & & & \\

TERMINATOR~\cite{nagle2026terminator}
& $\checkmark$ & $\checkmark$ & & $\triangle$ & & $\triangle$ & $\triangle$ & & \\

SABER~\cite{cheng2026saberstabilityawareearlyexit}
& $\checkmark$ & $\checkmark$ & & & & & $\checkmark$ & & \\

NEAT~\cite{liu-etal-2026-neat}
& $\checkmark$ & $\checkmark$ & & & & & $\triangle$ & & \\

ASAG~\cite{li2026stopreasoningwonthelp}
& $\checkmark$ & $\checkmark$ & & $\checkmark$ & & $\checkmark$ & $\checkmark$ & & \\

\midrule

\textbf{Ours}
& $\checkmark$ & $\checkmark$ & $\checkmark$ & $\checkmark$ & $\checkmark$
& $\checkmark$ & $\checkmark$ & $\checkmark$ & $\checkmark$ \\
\bottomrule
\end{tabular}%
\vspace{2pt}
\begin{minipage}{\textwidth}
\scriptsize
\textit{Notes.}
Math includes GSM8K, MATH and AIME; Sci. QA includes GPQA-Diamond.
Gen. QA covers broad-domain question answering; Complex Code requires
competitive-programming evaluation such as LiveCodeBench.
Fact. requires a dedicated factuality or hallucination benchmark, such as
TruthfulQA or HaluEval.
Oracle means a ground-truth-informed hindsight analysis of correct intermediate
answers; budget search or steps-to-solution estimates count only as related evidence.
Exit Dynamics means empirical correctness or answer changes across reasoning positions.
Post-stop Audit examines reasoning after forced termination, rather than hesitation
markers in provisional answers.
CoT Quality requires a manual or automated evaluation of the retained chain.
\end{minipage}
\end{table}

\section{Benchmark Design}

\subsection{Evaluation Methodology}

\begin{figure*}[t]
    \centering
    \resizebox{\textwidth}{!}{\definecolor{colRight}{HTML}{4FA35E}%
\definecolor{colFlowBlue}{HTML}{BFDCEB}%
\definecolor{colFlowPink}{HTML}{F2C4D3}%
\definecolor{colBoxFill}{HTML}{FBEEF3}%
\definecolor{colTokBlue}{HTML}{D3E5F3}%
\definecolor{colTokBlueEdge}{HTML}{8FB3D9}%
\definecolor{colTokGreen}{HTML}{CDEBCF}%
\definecolor{colPurpleEdge}{HTML}{9A86C4}%
\definecolor{colFire}{HTML}{F05A28}%
\definecolor{colIce}{HTML}{6CC3E8}%
\definecolor{colGold}{HTML}{E0A800}%
\definecolor{colRed}{HTML}{D94452}%
\definecolor{colInk}{HTML}{222222}%
\begin{tikzpicture}[
    font=\sffamily\scriptsize,
    group/.style={draw=black!45, dashed, rounded corners=14pt, line width=0.7pt},
    gtitle/.style={font=\sffamily\bfseries\small, text=colInk},
    tall/.style={draw=colInk, fill=colBoxFill, rounded corners=5pt, line width=0.8pt},
    flatbox/.style={draw=colInk, fill=colBoxFill, rounded corners=4pt, line width=0.7pt, align=center, font=\sffamily\scriptsize, text=colInk},
    blueflow/.style={-{Triangle[length=5pt,width=7pt]}, line width=3.2pt, colFlowBlue},
    pinkflow/.style={-{Triangle[length=5pt,width=7pt]}, line width=3.2pt, colFlowPink},
    thinblue/.style={-{Triangle[length=4pt,width=5pt]}, line width=2pt, colFlowBlue},
    thinpink/.style={-{Triangle[length=4pt,width=5pt]}, line width=2pt, colFlowPink},
    sq/.style={rounded corners=1.5pt, minimum size=0.26cm, inner sep=0pt},
    outsym/.style={font=\sffamily\bfseries\itshape\small, text=colInk},
]

\newcommand{\fire}{\textcolor{colFire}{\faFire}}
\newcommand{\ice}{\textcolor{colIce}{\faSnowflake}}

\newcommand{\tokcolumn}[6]{%
    \draw[#4, fill=#5, rounded corners=3pt, line width=0.7pt] (#1-0.27,#2-0.12) rectangle (#1+0.27,#2+0.36*#3-0.11);
    \foreach \k in {1,...,#3} {
        \pgfmathtruncatemacro{\kk}{\k}
        \ifnum\kk>#6
            \node[sq, fill=white, draw=black!15] at (#1,#2+0.36*\k-0.36+0.06) {};
        \else
            \node[sq, fill=#4!45] at (#1,#2+0.36*\k-0.36+0.06) {};
        \fi
    }
}

\node[gtitle] at (4.675,6.15) {(I) Generate Once};
\draw[group] (2.75,1.2) rectangle (6.6,5.8);

\node[gtitle] at (1.15,5.35) {Input};
\draw[fill=white, draw=colInk, line width=0.7pt] (0.6,3.25) -- (1.5,3.25) -- (1.75,3.5) -- (1.75,4.75) -- (0.6,4.75) -- cycle;
\draw[colInk, line width=0.7pt] (1.5,3.25) -- (1.5,3.5) -- (1.75,3.5);
\foreach \l/\c in {4.5/colTokBlueEdge,4.27/colTokBlueEdge,4.04/colPurpleEdge,3.81/colTokBlueEdge,3.58/colPurpleEdge} {
    \draw[\c, line width=1pt, line cap=round] (0.78,\l) -- (1.5,\l);
}
\node at (1.175,2.95) {question};
\node[font=\sffamily\tiny, text=black!55] at (1.175,2.62) {math, code, factuality};

\draw[blueflow] (1.9,4.0) -- (3.0,4.0);

\draw[tall] (3.1,1.65) rectangle (3.9,5.35);
\node[rotate=-90, font=\sffamily\small, text=colInk] at (3.5,3.6) {Reasoning Model};
\node at (3.5,1.95) {\ice};

\draw[blueflow] (4.0,3.5) -- (4.55,3.5);
\tokcolumn{4.95}{2.2}{7}{colTokBlueEdge}{colTokBlue}{6}
\node[sq, fill=colTokGreen, draw=colTokBlueEdge] at (4.95,2.2+0.36*6+0.06) {};
\node[align=center] at (4.95,1.75) {full\\[-1pt]trajectory};

\draw[blueflow] (5.33,3.5) -- (5.62,3.5);
\fill[colBoxFill] (5.59,2.35) rectangle (6.25,4.65);
\fill[colBoxFill] (5.92,2.35) ellipse (0.33 and 0.11);
\draw[colInk, line width=0.7pt] (5.59,4.65) -- (5.59,2.35) arc[start angle=180, end angle=360, x radius=0.33, y radius=0.11] -- (6.25,4.65);
\draw[colInk, line width=0.7pt, fill=colBoxFill] (5.92,4.65) ellipse (0.33 and 0.11);
\node[rotate=-90, text=colInk] at (5.92,3.45) {Trajectory Store};

\draw[blueflow] (6.25,3.5) -- (7.4,3.5);

\begin{scope}[xshift=0.25cm]
\node[gtitle] at (9.6,6.15) {(II) Replay \& Verify};
\draw[group] (6.75,1.2) rectangle (12.45,5.8);
\tokcolumn{7.5}{2.2}{8}{colTokBlueEdge}{colTokBlue}{5}
\draw[colRed, line width=1.1pt] (7.2,2.2+0.36*5-0.12) -- (7.8,2.2+0.36*5-0.12);
\node[font=\sffamily\tiny, text=colRed, anchor=east] at (7.2,2.2+0.36*5-0.12) {$t$};
\node at (7.5,1.88) {prefix};

\draw[blueflow] (7.85,3.5) -- (8.35,3.5);
\draw[tall] (8.45,2.05) rectangle (9.15,4.95);
\node[rotate=-90, font=\sffamily\small, text=colInk] at (8.8,3.35) {Shared Probe};
\node[font=\tiny\ttfamily, text=colRed, fill=white, draw=colRed!60, rounded corners=1pt, inner sep=1pt] at (8.8,4.7) {</think>};

\node[flatbox, minimum width=1.85cm, minimum height=0.62cm] (vs) at (10.45,4.6) {State-based \fire};
\node[flatbox, minimum width=1.85cm, minimum height=0.62cm] (vf) at (10.45,3.5) {Answer-free};
\node[flatbox, minimum width=1.85cm, minimum height=0.62cm] (va) at (10.45,2.4) {Answer-based};
\draw[thinblue] (9.2,3.5) -- (vf.west);
\draw[thinblue] (9.2,3.1) -- (9.4,3.1) -- (9.4,2.4) -- (va.west);
\draw[thinblue] (7.5,5.15) -- (7.5,5.4) -- (9.4,5.4) -- (9.4,4.6) -- (vs.west);
\node[font=\sffamily\tiny, text=colTokBlueEdge!70!black, fill=white, inner sep=1pt] at (8.45,5.4) {hidden states};

\node[circle, draw=colInk, fill=white, line width=0.8pt, minimum size=0.62cm, inner sep=0pt, font=\sffamily\tiny\bfseries] (dec) at (11.95,3.5) {Stop?};
\draw[thinpink] (vs.east) -- (11.95,4.6) -- (dec.north);
\draw[thinpink] (vf.east) -- (dec.west);
\draw[thinpink] (va.east) -- (11.95,2.4) -- (dec.south);

\draw[pinkflow] (11.95,2.15) -- (11.95,1.5) -- (7.5,1.5) -- (7.5,1.65);
\node[text=colRed!80!black, fill=white, inner sep=1.5pt] at (9.75,1.5) {continue to next checkpoint};

\draw[pinkflow] (12.28,3.5) -- (13.1,3.5);
\node[text=colInk] at (12.7,3.82) {stop};
\draw[tall] (13.2,1.65) rectangle (13.95,5.35);
\node[rotate=-90, font=\sffamily\small, text=colInk] at (13.575,3.5) {Final Answer};

\draw[blueflow] (14.05,4.2) -- (14.4,4.2);
\draw[black!70, line width=0.6pt, -{Stealth[length=3pt]}] (14.55,3.35) -- (14.55,5.15);
\draw[black!70, line width=0.6pt, -{Stealth[length=3pt]}] (14.55,3.35) -- (15.6,3.35);
\draw[black!45, dashed, line width=0.6pt] (14.65,3.5) .. controls (14.95,4.45) and (15.25,4.55) .. (15.55,4.6);
\fill[colTokBlueEdge!80!black] (15.05,4.0) circle (2pt);
\node[star, star points=5, fill=colGold, inner sep=0pt, minimum size=7pt] at (14.9,4.95) {};
\node[font=\sffamily\tiny, text=colGold!80!black, anchor=west, inner sep=1pt] at (15.03,4.95) {oracle};
\node[font=\sffamily\tiny, text=colTokBlueEdge!60!black, anchor=north, inner sep=1pt] at (15.05,3.9) {method};
\node[font=\sffamily\tiny, text=black!50, anchor=south west, inner sep=1pt] at (15.5,4.6) {fixed budgets};
\node[font=\sffamily\tiny, text=black!60, below=1pt] at (15.08,3.35) {tokens};
\node[font=\sffamily\tiny, text=black!60, rotate=90] at (14.42,3.75) {acc.};
\node[text=colInk] at (15.08,2.75) {vs. references};

\draw[pinkflow] (14.05,2.2) -- (14.85,2.2);
\node[outsym, font=\sffamily\bfseries\scriptsize] at (15.85,2.2) {Correct? \textcolor{colRight}{\faCheck}\,/\,\textcolor{colRed}{\faTimes}};
\draw[thinblue] (15.55,4.05) -- (15.8,4.05);
\node[outsym, anchor=west, font=\sffamily\bfseries\scriptsize, inner xsep=0pt] at (15.88,4.22) {Efficiency};
\node[outsym, anchor=west, font=\sffamily\bfseries\scriptsize, inner xsep=0pt] (cost) at (15.88,3.88) {Token cost};
\end{scope}

\draw[group] let \p1=(cost.east) in (13.1,1.2) rectangle (\x1+0.35cm,5.8);
\path let \p1=(cost.east) in node[gtitle] at ({(13.1cm+\x1+0.35cm)/2},6.15) {(III) Exit \& Evaluate};

\draw[fill=white, draw=colInk, line width=0.9pt, rounded corners=7pt] (0.2,0.55) rectangle (2.15,1.05);
\node at (1.175,0.8) {gold answer};
\draw[pinkflow] (2.2,0.8) -- (16.1,0.8) -- (16.1,1.95);

\draw[group, rounded corners=6pt] (3.45,-0.25) rectangle (14.35,0.4);
\node at (8.9,0.075)
    {\tikz[baseline=-3pt]\node[sq, fill=colTokBlueEdge!45]{};\, Prefix token
     \quad \tikz[baseline=-3pt]\node[sq, fill=white, draw=black!25]{};\, Hidden token
     \quad \tikz[baseline=-3pt]\node[sq, fill=colTokGreen, draw=colTokBlueEdge]{};\, Full answer
     \quad \fire\, Trained verifier
     \quad \ice\, Frozen
     \quad \tikz[baseline=-3pt]\node[star, star points=5, fill=colGold, inner sep=0pt, minimum size=6pt]{};\, Oracle};

\end{tikzpicture}%
}
    \caption{\textbf{Overview of the replay-based benchmark}. (I) The frozen reasoning model answers each question once, and the full trajectory is stored. (II) The stored trajectory is replayed. At each checkpoint, the prefix followed by \texttt{</think>} yields one shared probe, and internal-state based, answer-free and answer-based verifiers read either the internal states or this probe. Each method's stopping rule then decides to stop or to continue to the next checkpoint. (III) A method that stops generates its final answer, which is checked against the gold answer and compared with reference policies in accuracy and total token cost.}
    \label{fig:EE}
\end{figure*}
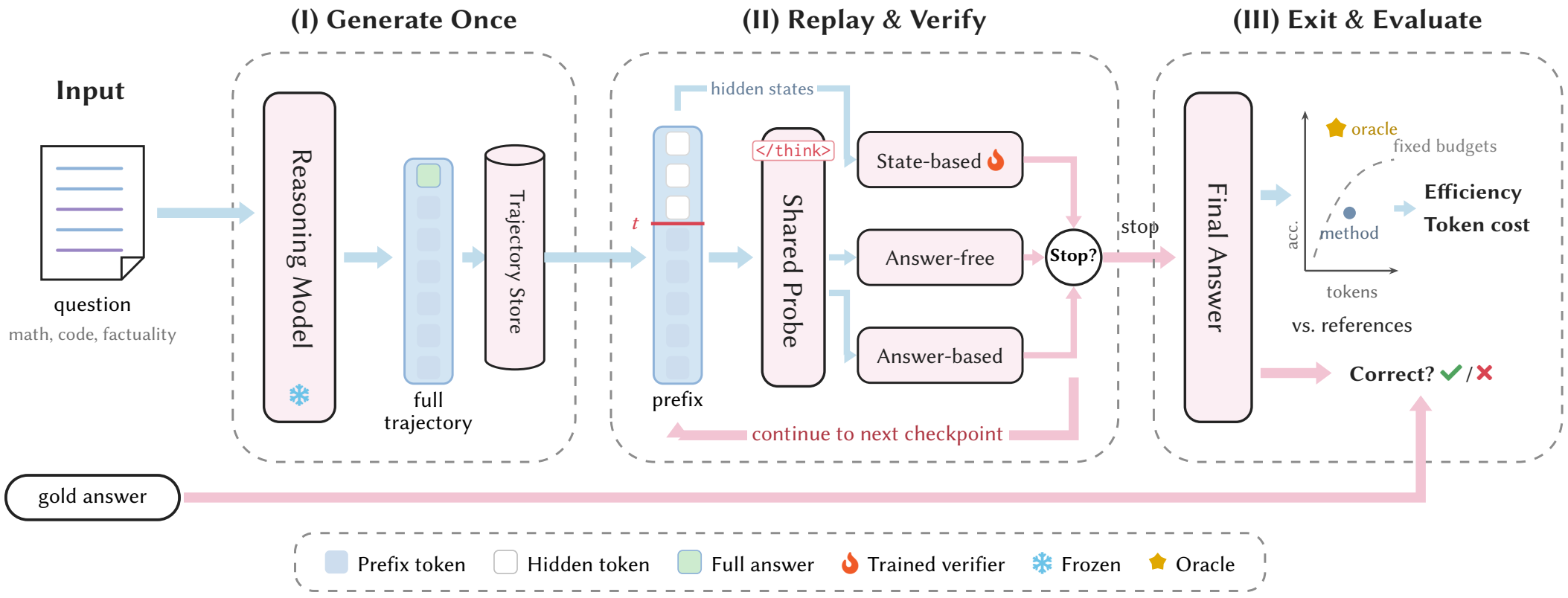

Evaluating many early-stopping methods independently can be expensive, because each method may require a separate model roll-out. To reduce this evaluation cost, we use a replay-based protocol, applied for both \textit{Training-free} and \textit{Training-based} methods. For each input example, we first generate one full reasoning trace from the model and store the generated tokens together with available metadata. We then replay this token sequence and evaluate all early-stopping methods on the same observed prefix stream. 
Let $\mathcal{D}$ be the evaluation dataset and let $M$ be the reasoning model. We denote the set of early-stopping methods by
\[
\mathcal{I}={1,2,\ldots,n}.
\]
Each method $m\in\mathcal{I}$ is defined by
\[
(S_m,\pi_m,G_m, \tau,v_m),
\]
where $v_m$ is the number of probe tokens required by the verifier. Since all methods share the relatively same $\tau_m$, we share the same $\tau$ across all methods. When $v_m=0$, method $m$ does not use probe answers. For each example $(x,g)\in\mathcal{D}$, we first generate a full reasoning trajectory:
\[
(r_{T},y^{\mathrm{full}})\sim M(x),
\]
During replay, each early-stopping method observes only the prefix $r_{t}$ at time $t$. We let $\mathcal{M}_t$ as collection of methods that is being evaluated at $t$, and the temporal answer for verification is

\[
y_t^{\mathrm{probe}}
\sim M\left(
\cdot \mid x,\; r_t \circ \tau;\;
\mathrm{max\_tokens}=v^{\max}
\right),
\qquad
v_{\max} =
            \max_{m \in \mathcal{M}_t} v_m
\]

 Method $m$  only receives and evaluates the first $v_m$ probe answer tokens with $\pi_m$:
\[
u_{m,t}
\leftarrow
\pi_m
\left(
x,
r_{t},
\mathbf{y}_{t,v}^{\mathrm{probe}}
\right).
\]

When method $m$ decides to stop at time $t$, we append $\tau$ to the reasoning prefix and generate its early-stopping answer:
\[
y_m\sim M(x,r_{t}\circ\tau).
\]
The early-stopping answer $y_m$ is generated separately from the shared probe answers because probe answer is not a completed answer. $y_m$ is then evaluated by the scoring function $V$. After method $m$ stops, it is removed from the active method set for that sample.

Under identical prefixes, decoding settings, and verifier outputs, the replay protocol produces the same stopping decisions as online execution, provided the method does not modify the ongoing reasoning trajectory.

\begin{algorithm}[t]
\small
\caption{Replay-Based Evaluation for Early Exit Methods}
\label{alg:replay_eval}
\begin{algorithmic}[1]


\State Initialize result set $\mathcal{R} \gets \emptyset$

\For{each example $(x,g) \in \mathcal{D}$}
    \State $(r_{1:T}, y^{\mathrm{full}}) \sim M(x), \quad\mathcal{A} \gets \mathcal{I}$

    \For{$t = 1$ to $T$}

        \State $\mathcal{M}_t \gets
        \{m \in \mathcal{A} : S_m(x,r_{1:t}) = 1\}$

        \If{$\mathcal{M}_t \neq \emptyset$}
            \State $v_{\max} \gets
            \max_{m \in \mathcal{M}_t} v_m$

            \State $\mathbf{y}_{t,v_{\max}}^{\mathrm{probe}}
            \sim M(x, r_{t} \circ \tau)$

            \For{each method $m \in \mathcal{M}_t$}
                \State $u_{m,t} \gets
                \pi_m
                \left(
                    x,
                    r_{1:t},
                    \mathbf{y}_{t,v_m}^{\mathrm{probe}}
                \right)$

                \If{$G_m(u_{m,t}, r_{t}) = 1$}
                    \State $y_m \sim M(x, r_{t} \circ \tau)$

                    \State $a_m \gets
                    V(y_m, g)$

                    \State Add $(x,m,t,y_m,a_m)$ to $\mathcal{R}$

                    \State Remove $m$ from $\mathcal{A}$
                \EndIf
            \EndFor
        \EndIf

    \EndFor

    \State $a_{full} \gets
        V(y^{\mathrm{full}}, g)$

    \For{each remaining method $m \in \mathcal{A}$}

        \State Add
        $(x,m,T,y^{\mathrm{full}},a_{full})$
        to $\mathcal{R}$
    \EndFor
\EndFor

\State \textbf{return} $\mathcal{R}$

\end{algorithmic}
\end{algorithm}

\subsection{Random stop positions}

In addition to evaluating the exit positions selected by each method, we perform a randomized exit analysis to reduce bias introduced by the stopping policies themselves and to study whether non-monotonic reasoning behavior is a general property of the reasoning trajectory rather than an artifact of a particular method.

\begin{figure*}[t]
    \centering
    \includegraphics[width=0.9\textwidth]{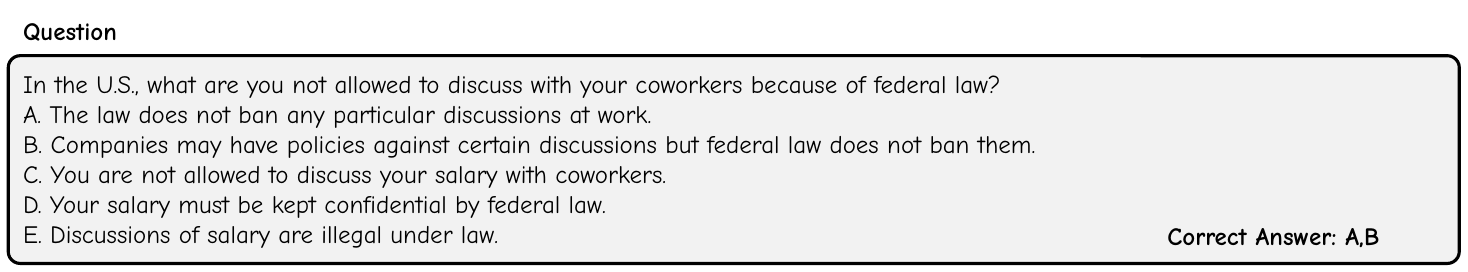}
    \par\medskip
    \includegraphics[width=0.9\textwidth]{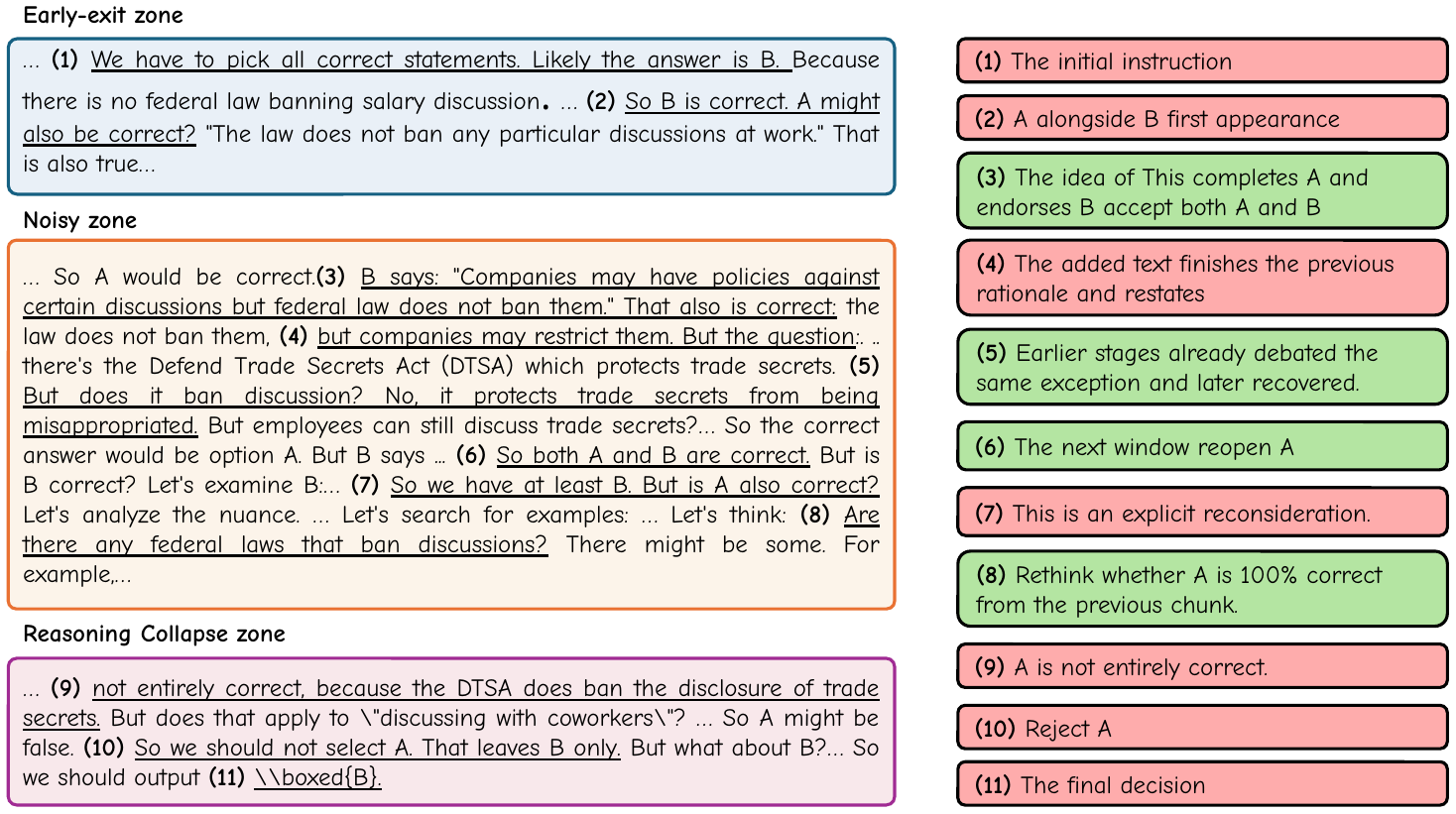}
    \caption{Random exit accuracy for TruthfulQA question 229
    using gpt-oss-20b. Colors indicate accuracy,
    with \textcolor{red!75!black}{red (wrong answer)}
    to \textcolor{green!60!black}{green (right answer)}.}
    \label{fig:random_text}
\end{figure*}

For each selected question, we sample exit positions
along the recorded reasoning trajectory and generate
eight independent answers at each sampled position
using the same decoding configuration. The sampled questions are distributed across the benchmarks as follows: all from AIME26 and 128-196 from each MATH500, MMLU-Pro, TruthfulQA, and LiveCodeBench. An example of the randomized exit evaluation is shown in Figure~\ref{fig:random_text}.

\subsection{Adjustment from original implementation}


\paragraph{Adjustment to Preserve KV Cache optimization}
For methods that score the output like DTSR, the original implementation ask for confidence scoring via a new chat template, which invalidates the KV cache from previous generation. To prevent this, we use its variant DTSR-1 \cite{xiang-etal-2026-thinking} that adding the scoring prompt after reasoning chunk $r_t$,

\paragraph{Adjustment for Coding benchmark}

\begin{figure}[htbp]
    \centering

    \begin{subfigure}{0.49\textwidth}
        \centering
        \includegraphics[width=\linewidth]{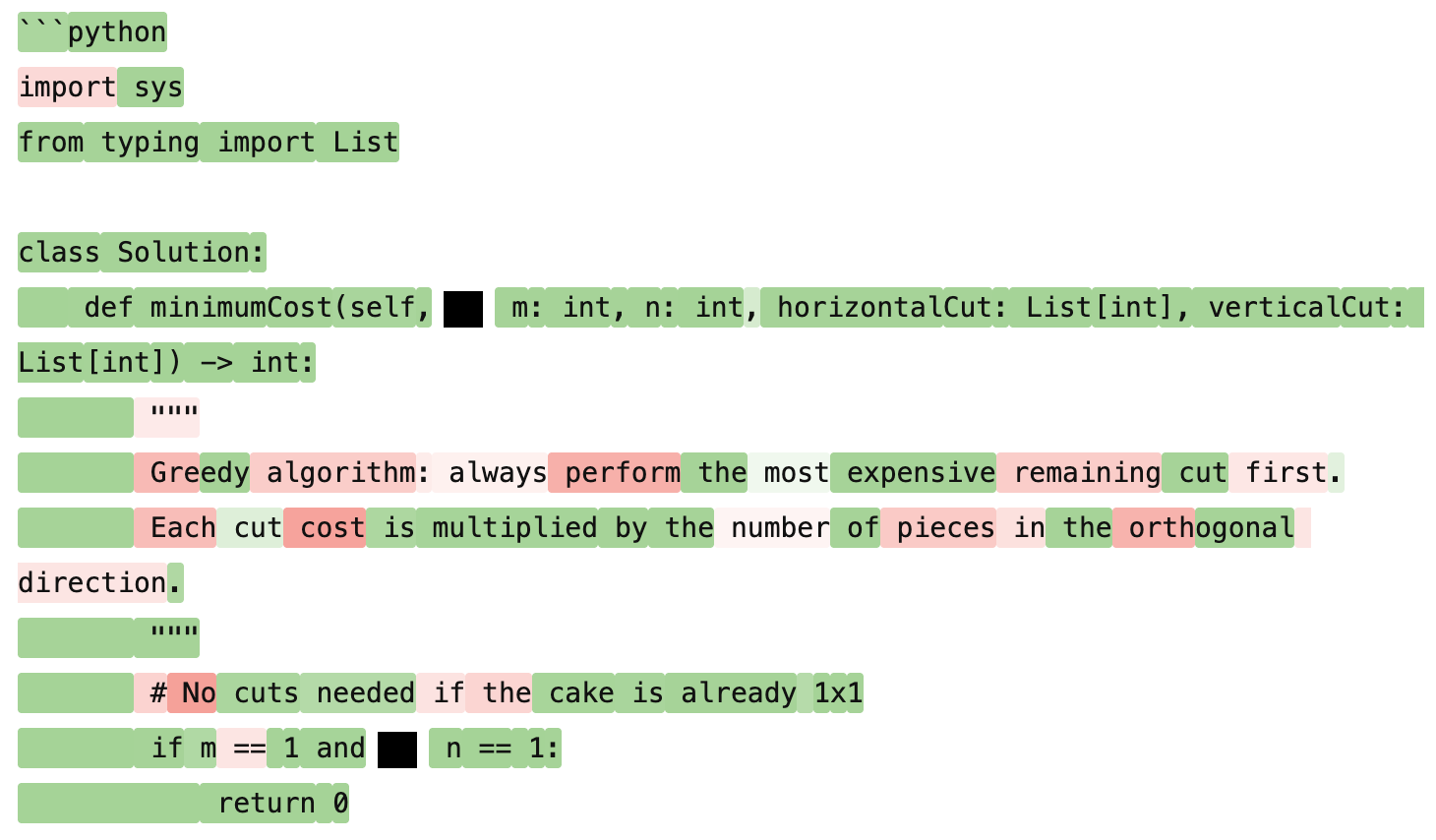}
        \subcaption{Probability score for DEER}
        \label{fig:chart1}
    \end{subfigure}
    \hfill
    \begin{subfigure}{0.49\textwidth}
        \centering
        \includegraphics[width=\linewidth]{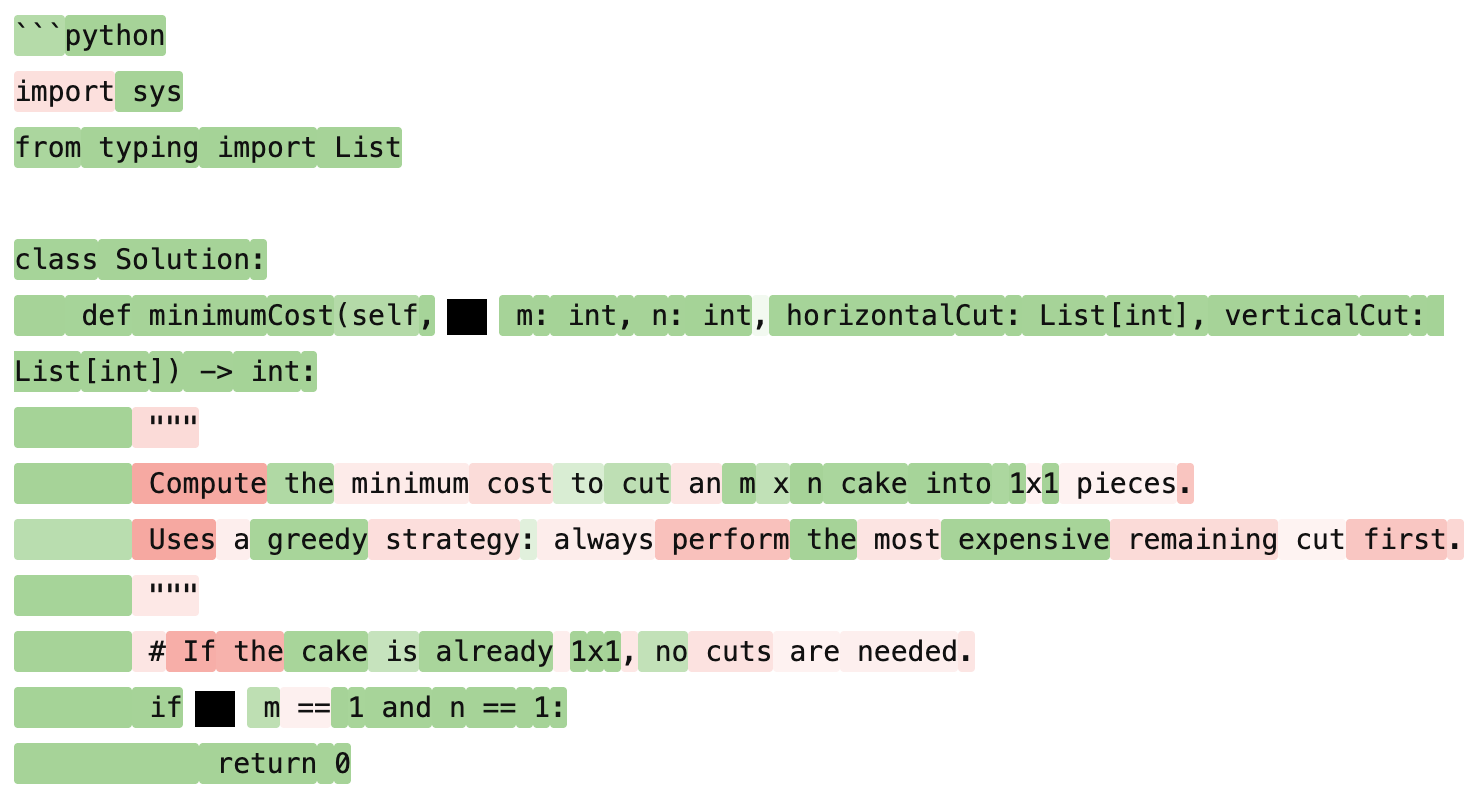}
        \subcaption{Entropy score for EntroCut}
        \label{fig:chart2}
    \end{subfigure}

    \caption{Scoring for LiveCodeBench question 585: minimum-cost-for-cutting-cake-i (substring only). The two $\blacksquare$ markers indicate token limits of 20 and 100, respectively. \textcolor{green!60!black}{Green} highlights tokens that exceed the corresponding thresholds. The first 20 tokens of both solutions consist of imports and function declarations, while the scores only fluctuate in later sections}
    \label{fig:comparison}
\end{figure}

Answer-free methods often temporarily stop and request an immediate final answer. This works well for mathematics and multiple-choice tasks, where answers are short, and first few tokens contain the entire option. Coding tasks require much longer output, and the first few tokens are normally solution template (Figure~\ref{fig:comparison}).  We therefore increase the budget for probe answers by 5 times and reduce the scale of threshold score for methods by 0.99 to ensure the probe answers have more information while not being too harsh for extra steps.

Since answer-based methods require complete provisional code solutions during verification, which substantially increases the cost, we exclude them from the main coding comparison.

\section{Experimental Setup and Evaluation Metrics}
\label{sec:evaluation}

\subsection{Experimental Setup}
\label{sec:setup}

\paragraph{Models.}
We evaluate Gemma 4 E4B \citep{gemmateam2026gemma4technicalreport}, Qwen3.5-4B and Qwen3.5-9B \citep{qwen3.5}, R1-Qwen3-8B \citep{deepseekai2025deepseekr1incentivizingreasoningcapability}, and gpt-oss-20b at high reasoning effort \citep{openai2025gptoss120bgptoss20bmodel}.

\paragraph{Datasets.}

\begin{table}[t]
    \centering
    \caption{Summary of evaluation datasets.}
    \label{tab:datasets}
    \footnotesize
\setlength{\tabcolsep}{2.5pt}
\renewcommand{\arraystretch}{0.95}
    \begin{tabular}{lrl}
        \toprule
        \textbf{Dataset} & \textbf{Samples} & \textbf{Format} \\
        \midrule
        AIME26 & 30 & Free-text \\
        MATH500 & 500 & Free-text \\
        GSM8K & 1,319 & Free-text \\
        GPQA-Diamond & 198 & MCQ \\
        MMLU-Pro (subset) & 2,406 & MCQ \\
        \midrule
        \makecell{LiveCodeBench v6 Lite} & 1,055 & Test cases \\
        \midrule
        HaluEval (General) & 4,507 & Binary \\
        TruthfulQA & 817 & MCQ \\
        \bottomrule
    \end{tabular}
\end{table}

We evaluate on eight benchmarks spanning: mathematical and scientific reasoning include AIME26 \cite{aime26}, MATH500 \cite{lightman2023lets}, GSM8K \cite{cobbe2021trainingverifierssolvemath}, GPQA-Diamond \cite{rein2023gpqagraduatelevelgoogleproofqa} and a subset of MMLU-Pro \cite{3737916.3740934}; code generation with LiveCodeBench v6 Lite \cite{jain2024livecodebench}; and factuality or hallucination evaluation with TruthfulQA \cite{lin2022truthfulqameasuringmodelsmimic} and the General set of HaluEval \cite{li2023haluevallargescalehallucinationevaluation}. We conducted 3 runs and collected the macro average for AIME26 and GPQA-Diamond and single pass for the rest.

For training methods, we use MATH \cite{hendrycksmath2021} without the duplication in MATH500 and AIME 1983-2024 \cite{veeraboina2024aime} with the same curation pipeline of using greedy to find the nearest valid exit position as the breakpoint and label the hidden position accordingly.

\paragraph{Early-stopping methods.}
We evaluate methods from the early-stopping categories described in Section~\ref{subsec:taxonomy}. Training-based methods include TERMINATOR and Thought Calibration that need to be retrained. Training-free methods include two groups: answer-free method: Statistical Early Exit, DEER, CoDE-Stop, EntroCut, DTSR-1 and answer-based method: Answer Convergence with stopping-rule variant, DynaSor, and PUMA. Even though PUMA was originally a training-based method, we reuse the original checkpoint. Detailed configuration can be found in Appendix~\ref{app:individual-replay-methods}.

\paragraph{Hybrid policies.}
We additionally evaluate a rule-based policy that combines multiple stopping methods. At checkpoint $\theta_k$, let $\Gamma(\theta_k)$ denote the set of active methods. For exit-decision agreement, let $e_m(\theta_k)\in\{0,1\}$ denote whether method $m$ recommends stopping. We define
\[
R_{\mathrm{exit}}(\theta_k)
=
\frac{1}{|\Gamma(\theta_k)|}
\sum_{m\in\Gamma(\theta_k)}
e_m(\theta_k),
\]
and exit when
$
R_{\mathrm{exit}}(\theta_k)\geq\rho.
$

\paragraph{Generation settings and token budgets.}
The reasoning trace is capped at $2^{15}$ tokens for all models. Probe generations used by verifier-based methods are capped at $100$ tokens for code and $20$ tokens for others, while final-answer generation is capped at $2{,}048$ tokens. All use temperature of $0.6$. We evaluate fixed-budget reasoning using a predefined budget set: 

\[
\mathcal{B} = \{2^{9},2^{10},3.2^{9},2^{11},3.2^{10},2^{12},3.2^{11},2^{13},3.2^{12},2^{14},3.2^{13},2^{15}\}.
\]



\subsection{Evaluation References and Cost Accounting}
\label{subsec:evaluation_references}

We compare methods with no thinking, natural full reasoning, fixed-budget forcing for $b\in\mathcal B$, and a \emph{discrete-budget oracle}.
For each question $i$, the oracle selects the smallest evaluated reasoning budget that produces a correct answer:
\[
b_i^{\star}
=
\min
\left\{
b\in\mathcal{B}:
V(y_i^{(b)},g_i)=1
\right\},
\]
where $g_i$ is the ground-truth target. If no evaluated budget produces a correct answer, we set
$
b_i^{\star}=\infty.
$
A hypothetical all-position oracle would consider every token position under the same stopping intervention. However, the empirical oracle searches only
 the evaluated budget set $\mathcal{B}$ and selects the smallest successful reasoning
 budget in that set. Throughout the
 experiments, “oracle” refers to this discrete-budget reference.

We also report total generated-token cost from reasoning tokens, final-answer tokens, verifier or sufficiency-probe tokens, and temporary-answer tokens as:
\[
C_{m,i}
=
C^{\mathrm{reason}}_{m,i}
+
C^{\mathrm{answer}}_{m,i}
+
C^{\mathrm{probe}}_{m,i}
+
C^{\mathrm{temp}}_{m,i},
\]

\subsection{Accuracy and Oracle-Relative Efficiency}
\label{subsec:efficiency_metrics}

\paragraph{Accuracy targets.}
Let $\bar a_0$ and $\bar a_H$ denote the mean accuracies of the
no-thinking baseline and the hindsight oracle, respectively. We construct normalized accuracy targets
$q\in(0,1]$, and the fraction of the available accuracy gain
required above the no-thinking baseline as 
\[
a(q)=\bar a_0+q\Delta a,\qquad \Delta a = \bar a_H-\bar a_0.
\]

\paragraph{Cumulative success-cost curve.}
For each method $m$
and question $i$, let $p_{m,i}\in[0,1]$ denote the fraction of
repeated runs answered correctly, and let $c_{m,i}$ denote the
mean total cost. We define the empirical cumulative success-cost curve as
\[
F_m(c)
=
\frac{1}{N}
\sum_{i=1}^{N}
p_{m,i}\,
\mathbb{I}[c_{m,i}\leq c].
\]

This curve measures the fraction of all evaluation questions
successfully answered with observed mean token cost no greater
than $c$. As $c$ increases, $F_m(c)$ is non-decreasing and
approaches the method's overall accuracy. For a normalized accuracy target $q$, the corresponding token
requirement is defined by the generalized inverse
\[
C_m(q)
=
\inf\left\{
c\geq0:
F_m(c)\geq a(q)
\right\}.
\]
We set $C_m(q)=+\infty$ if method $m$ cannot attain the target.
The token requirement for oracle solution is $C_H(q)$.

\paragraph{Continuous formulation.}
For each accuracy target, we define oracle-relative token
efficiency as
\[
e_m(q)=
\begin{cases}
\displaystyle
\min\left(
1,\frac{C_H(q)}{C_m(q) + \epsilon}
\right),
& C_m(q)<+\infty,\\[8pt]
0,
& C_m(q)=+\infty.
\end{cases}
\]
The ratio is capped at one, so matching or outperforming the oracle's
token requirement receives full efficiency at that target.
Unattainable targets contribute zero. The continuous form of TITE is
\[
\boxed{
\mathrm{TITE}_m
=
\int_0^1 e_m(q)\,dq.
}
\]

\paragraph{Discrete implementation.}

In practice, we evaluate TITE over $1{,}000$ uniformly spaced
normalized targets,
\[
Q=\left\{
\frac{j}{1000}:j=1,\ldots,1000
\right\}.
\]

$C_m(q)$ is obtained by sorting
question-level mean token costs and selecting the smallest
observed cost threshold at which cumulative success reaches
$a(q)$. The oracle requirement $C_H(q)$ is computed using the
same procedure. TITE formula is
\[
\boxed{
\mathrm{TITE}
=
\frac{1}{|Q|}
\sum_{q\in Q} e_m(q).
}
\]

\paragraph{Why target integration?}

\begin{figure*}[ht]
    \centering
    \includegraphics[width=0.9\textwidth]{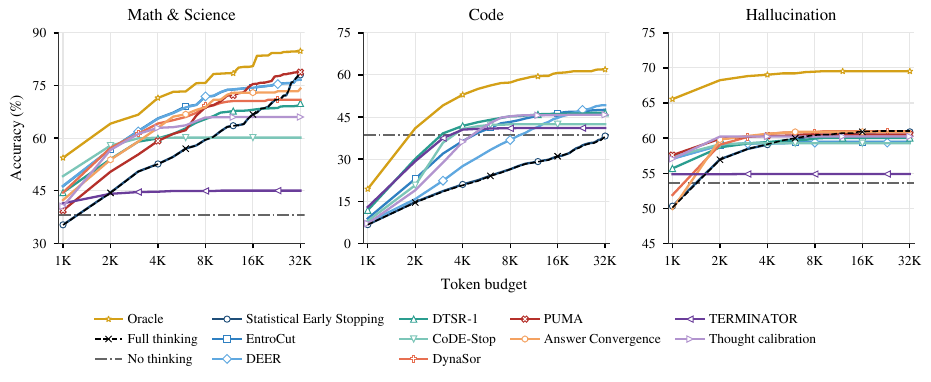}

    \caption{Token-accuracy figures of gpt-oss-20b}
    \label{fig:budget_accuracy}
\end{figure*}

A central difficulty in evaluating early stopping is that accuracy
and computational cost are competing objectives. Accuracy and mean generated-token cost capture two necessary aspects of early stopping, but they do not describe how successful answers are distributed across the costs observed. 
 While additional computation
improves accuracy in some settings, other methods exhibit
early saturation, with limited gains from larger budgets (Figure~\ref{fig:budget_accuracy}).
Therefore, comparisons at a single budget do not
fully characterize performance. A common alternative is to combine the two averages through a linear utility function $S_m(\lambda)$ \cite{ehab2026osprunerpruningchainsofthoughtreasoning} or ratio-based metric $R_m$ \cite{yan2026entrocutentropyguidedadaptivetruncation}:
\[
S_m(\lambda)
=
\bar a_m-\lambda\bar C_m,
\qquad
R_m=
\frac{\bar a_m-\bar a_0}{\bar C_m-\bar C_0},
\]

TITE instead evaluates a sequence of accuracy requirements using
the full empirical cumulative success-cost curve. This permits
methods with identical aggregate accuracy and token consumption
to receive different scores when their successful answers occur
at different observed token costs. Oracle normalization further
provides a target-specific reference for computational
efficiency, while penalizing methods that fail to achieve sufficient accuracy and avoids selecting a global token-budget integration range.

\subsection{Continuation Stability and Monotonicity validation}
\label{subsec:preservation_recovery}

\begin{figure*}[ht]
    \centering
    \includegraphics[width=0.9\textwidth]{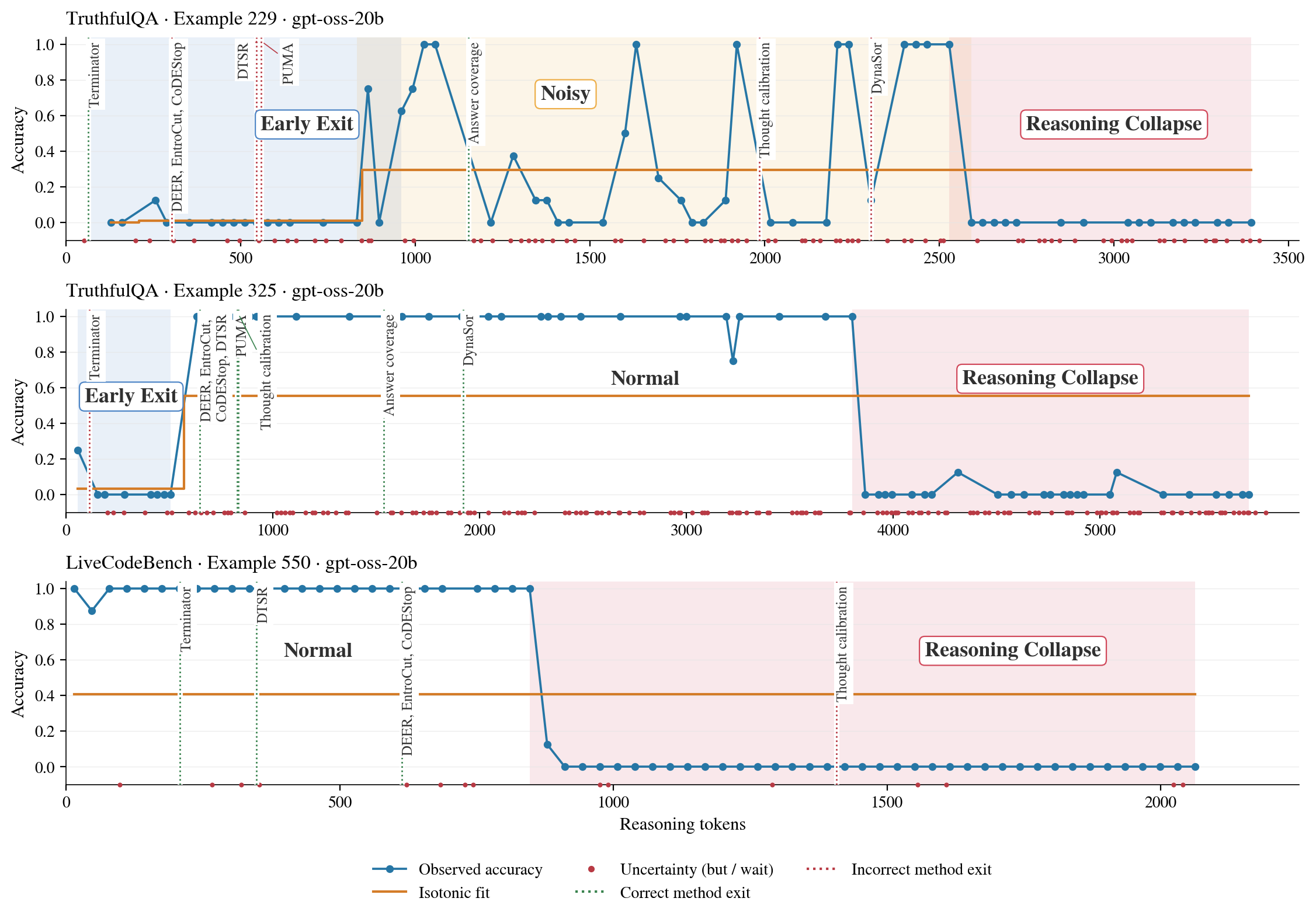}

    \caption{Visualization of random exit position and exit traps.}
    \label{fig:trace}
\end{figure*}

Aggregate early-stopping accuracy does not show how stable correctness is along a reasoning trajectory. In particular, a correct answer at an earlier point may be followed by an incorrect answer after additional reasoning. 

We evaluate this behavior in two stages. First, the \emph{continuation reasoning-error rate} measures how often observed later attempts fail after a correct earlier attempt. Second, the \emph{random-exit monotonicity test} checks whether these reversals are larger than what can be explained by stochastic decoding alone.

\paragraph{Continuation reasoning-error rate.}
Let $\alpha=(r,m,t)$ denote a correct anchor answer produced by method $m$ on reasoning trajectory $r\in\mathcal{R}$ at position $t$. For this anchor, define the set of valid later attempts

\[
L_\alpha
=
\{j : r_j=r,\; m_j\neq m,\; t_j>t,\; j \text{ is valid}\}.
\]

Let $c_j=1$ if later attempt $j$ is correct and $c_j=0$ otherwise. For every correct anchor with at least one valid later attempt, its continuation error is

\[
{CER}_\alpha
=
\frac{1}{|L_\alpha|}
\sum_{j\in L_\alpha}(1-c_j),
\qquad
CER
=
\frac{1}{\mathcal{R}}
\sum_{\alpha=1}{CER}_\alpha.
\]

\paragraph{Random-exit monotonicity validation.}
The first test relies on exit position of methods, which can be biased due to the algorithm. This test generalize the exit position by selecting random exit position to show that regardless of methods, early-stopping itself show the non-monotonic behavior. For question $i$, let

\[
t_{i1}<t_{i2}<\cdots<t_{iK_i}
\]

be its valid sampled exits, and let $p_{ik}$ denote the probability of producing a correct answer when exiting at $t_{ik}$. The monotone null hypothesis is

\[
H_{0,i}:
p_{i1}
\le
p_{i2}
\le
\cdots
\le
p_{iK_i}.
\]

At each exit, repeated generations provide a binomial estimate $\widehat p_{ik}$. We fit the best non-decreasing sequence $\widehat p^{\mathrm{iso}}_{ik}$ using binomial isotonic regression. For trajectory $i$, the normalized likelihood discrepancy $d_i$ and the average results $D_{\mathrm{overall}}$  are:

\[
d_i
=
\frac{2}{K_i}
\left[
\ell_i(\widehat{\mathbf p}_i)
-
\ell_i(\widehat{\mathbf p}^{\mathrm{iso}}_i)
\right],
\qquad
D_{\mathrm{overall}}
=
\frac{1}{N}\sum_{i=1}^{N} d_i.
\]

We additionally report an interpretable reversal-magnitude statistic. For question $i$,

\[
A_i
=
\frac{2}{K_i(K_i-1)}
\sum_{a<b}
max(
\widehat p_{ia}
-
\widehat p_{ib}
,0),
\]

This measures the average positive drop from an earlier exit to a later exit. Since some positive drops are expected from decoding noise, we report the excess reversal in magnitude $A_{\mathrm{excess}}$

\[
A_{\mathrm{excess}}
=
A_{\mathrm{obs}}
-
E_0[A^*],
\qquad
A_{\mathrm{obs}}=\frac{1}{N}\sum_{i=1}^{N}A_i
\]

where $E_0[A^*]$ is the expected value under the fitted monotone bootstrap.

\subsection{Output Reliability}
\label{subsec:output_reliability}

\paragraph{Post-stop reasoning continuation.}
We test whether the model continues reasoning after the stopping intervention using a lexical heuristic applied to the final-answer channel.  We distinguish primary markers, including explicit thinking delimiters and model-specific serialization
boundaries, from soft lexical markers. The latter consist of case-insensitive reasoning-continuation markers such as \texttt{wait}, \texttt{hmm} and \texttt{actually}. Let
\[
L_{m,i,r}
=
\operatorname{CountPostStopSignals}
\left(
y_{m,i,r}
\right)
\]
denote the number of detected continuation markers. The continuation rate for method $m$ is
\[
\operatorname{Leak}_m
=
\frac{1}{N_m}
\sum_{i,r}\ell_{m,i,r},
\quad
\ell_{m,i,r}
=
\mathbb{I}\left[L_{m,i,r}>0\right].
\]
We normalize each
stopping position by the length of its corresponding
full reasoning trace.


\paragraph{Retained reasoning-chain justification.}
A correct answer may still be produced from a reasoning prefix that does not contain enough information to justify the conclusion. We therefore use DeepSeek-V4.1-Flash \cite{deepseekai2026deepseekv41flash} as independent judge to score whether the retained reasoning supports the final answer. For each evaluated output, the judge receives the question, reference information available for evaluation, the retained reasoning prefix, and the model's final answer. It produces a justification score
$
J(r_{1:t})\in[10,100]
$
and a primary failure category.

The failure categories distinguish \textit{insufficient derivation}, \textit{unsupported assertions}, \textit{circular repetition}, contradiction with the final answer, and other failures. For each model, benchmark, and method, we stratify the evaluation by answer correctness and report
$
\bar J_m^{\,\mathrm{correct}},
\bar J_m^{\,\mathrm{incorrect}}.
$
We select $k=100$ samples per correctness class. We compare these values with the corresponding full-reasoning reference:
\[
\Delta J_m^{\,\mathrm{correct}}
=
\bar J_m^{\,\mathrm{correct}}
-
\bar J_{\mathrm{Thinking}}^{\,\mathrm{correct}},
\]
\[
\Delta J_m^{\,\mathrm{incorrect}}
=
\bar J_m^{\,\mathrm{incorrect}}
-
\bar J_{\mathrm{Thinking}}^{\,\mathrm{incorrect}}.
\]

A negative $\Delta J$ indicates that the early-stopping output retains weaker justification than full reasoning on the same model and benchmark. 

\section{Experiments Results}
\label{sec:results}

\subsection{RQ1: Efficiency, Generalization and Oracle Gap}
\label{subsec:rq1_efficiency}

\paragraph{Comparison with the discrete-budget oracle.}

\begin{table*}[t]
\centering
\scriptsize

\caption{Average accuracy and TITE on the Mathematical \& Science benchmarks. The best and second-best results in each column are
\textbf{bolded} and \underline{underlined}, respectively.}
\label{tab:acc-tite-math-qa}

\resizebox{\textwidth}{!}{%
\begin{tabular}{lcccccccccc}
\toprule

Method
& \multicolumn{2}{c}{Gemma4 E4B}
& \multicolumn{2}{c}{gpt-oss-20b}
& \multicolumn{2}{c}{Qwen3.5-4B}
& \multicolumn{2}{c}{Qwen3.5-9B}
& \multicolumn{2}{c}{R1-Qwen3-8B} \\

\cmidrule(lr){2-3}
\cmidrule(lr){4-5}
\cmidrule(lr){6-7}
\cmidrule(lr){8-9}
\cmidrule(lr){10-11}

&
Acc. & TITE
& Acc. & TITE
& Acc. & TITE
& Acc. & TITE
& Acc. & TITE \\

\midrule

Oracle
& \textbf{0.703} & --
& \textbf{0.894} & --
& \textbf{0.865} & --
& \textbf{0.891} & --
& \textbf{0.853} & -- \\

Full Thinking
& \underline{0.604} & --
& \underline{0.824} & --
& \underline{0.802} & --
& 0.821 & --
& \underline{0.766} & -- \\

No thinking
& 0.297 & --
& 0.381 & --
& 0.469 & --
& 0.513 & --
& 0.269 & -- \\

\midrule

Statistical EE
& 0.590 & \underline{0.342}
& \underline{0.824} & 0.191
& \underline{0.802} & 0.151
& \underline{0.841} & 0.176
& \underline{0.766} & 0.196 \\

EntroCut
& 0.574 & 0.333
& 0.773 & \underline{0.380}
& 0.700 & \underline{0.269}
& 0.741 & \textbf{0.207}
& 0.714 & \textbf{0.382} \\

DEER
& 0.573 & 0.331
& 0.773 & \textbf{0.381}
& 0.686 & 0.223
& 0.740 & 0.193
& 0.727 & \underline{0.362} \\

DTSR-1
& 0.586 & 0.336
& 0.704 & 0.281
& 0.529 & 0.126
& 0.539 & 0.052
& 0.523 & 0.282 \\

CoDE-Stop
& 0.571 & 0.333
& 0.601 & 0.254
& 0.532 & 0.118
& 0.590 & 0.087
& 0.508 & 0.300 \\

\midrule

DynaSor
& 0.534 & 0.317
& 0.709 & 0.286
& 0.519 & 0.079
& 0.599 & 0.094
& 0.628 & 0.294 \\

PUMA
& 0.563 & 0.325
& 0.799 & 0.278
& 0.651 & 0.108
& 0.784 & 0.165
& 0.738 & 0.293 \\

Answer Convergence
& 0.577 & 0.328
& 0.741 & 0.289
& 0.656 & 0.164
& 0.721 & 0.163
& 0.721 & 0.247 \\

\midrule

TERMINATOR
& 0.302 & 0.090
& 0.450 & 0.089
& 0.572 & 0.079
& 0.633 & 0.077
& 0.286 & 0.017 \\

Thought calibration
& 0.493 & 0.240
& 0.661 & 0.216
& 0.308 & 0.019
& 0.519 & 0.042
& 0.553 & 0.189 \\

\midrule

Exit agreement
& 0.458 & \textbf{0.406}
& 0.782 & 0.337
& 0.697 & \textbf{0.299}
& 0.747 & \underline{0.204}
& 0.706 & 0.341 \\

\bottomrule
\end{tabular}%
}
\end{table*}

\begin{table*}[t]
\centering
\scriptsize

\caption{Average accuracy and TITE on LiveCodeBench. The best and second-best results in each column are
\textbf{bolded} and \underline{underlined}, respectively.}
\label{tab:acc-tite-code}

\resizebox{\textwidth}{!}{%
\begin{tabular}{lcccccccccc}
\toprule

Method
& \multicolumn{2}{c}{Gemma4 E4B}
& \multicolumn{2}{c}{gpt-oss-20b}
& \multicolumn{2}{c}{Qwen3.5-4B}
& \multicolumn{2}{c}{Qwen3.5-9B}
& \multicolumn{2}{c}{R1-Qwen3-8B} \\

\cmidrule(lr){2-3}
\cmidrule(lr){4-5}
\cmidrule(lr){6-7}
\cmidrule(lr){8-9}
\cmidrule(lr){10-11}

&
Acc. & TITE
& Acc. & TITE
& Acc. & TITE
& Acc. & TITE
& Acc. & TITE \\

\midrule

Oracle
& \textbf{0.443} & --
& \textbf{0.624} & --
& \textbf{0.518} & --
& \textbf{0.624} & --
& \textbf{0.514} & -- \\

Full Thinking
& 0.415 & --
& 0.510 & --
& 0.424 & --
& 0.530 & --
& \underline{0.382} & -- \\

No thinking
& 0.278 & --
& 0.385 & --
& 0.254 & --
& 0.356 & --
& 0.191 & -- \\

\midrule

Statistical EE
& \underline{0.416} & \underline{0.303}
& 0.514 & 0.038
& \underline{0.426} & 0.052
& \underline{0.535} & \textbf{0.081}
& 0.379 & 0.048 \\

EntroCut
& 0.397 & 0.297
& 0.478 & 0.098
& 0.323 & 0.046
& 0.385 & 0.015
& 0.267 & 0.025 \\

DEER
& 0.413 & 0.286
& \underline{0.528} & 0.074
& 0.370 & 0.045
& 0.445 & 0.029
& 0.365 & 0.026 \\

DTSR-1
& 0.406 & 0.296
& 0.464 & \textbf{0.141}
& 0.290 & 0.025
& 0.342 & 0.000
& 0.247 & \underline{0.051} \\

CoDE-Stop
& 0.394 & 0.271
& 0.424 & 0.073
& 0.316 & \underline{0.057}
& 0.364 & 0.007
& 0.226 & \textbf{0.058} \\

\midrule

TERMINATOR
& 0.331 & 0.232
& 0.410 & 0.051
& 0.374 & 0.027
& 0.450 & 0.029
& 0.211 & 0.029 \\

Thought calibration
& 0.387 & \textbf{0.468}
& 0.458 & 0.104
& 0.323 & \textbf{0.074}
& 0.388 & 0.033
& 0.295 & 0.044 \\

\midrule

Exit agreement
& 0.411 & 0.282
& 0.499 & \underline{0.129}
& 0.324 & \textbf{0.074}
& 0.447 & \underline{0.047}
& 0.266 & 0.021 \\

\bottomrule
\end{tabular}%
}
\end{table*}

\begin{table*}[t]
\centering
\scriptsize

\caption{Average accuracy and TITE on the hallucination benchmarks. The best and second-best results in each column are
\textbf{bolded} and \underline{underlined}, respectively.}
\label{tab:acc-tite-hallucination}

\resizebox{\textwidth}{!}{%
\begin{tabular}{lcccccccccc}
\toprule

Method
& \multicolumn{2}{c}{Gemma4 E4B}
& \multicolumn{2}{c}{gpt-oss-20b}
& \multicolumn{2}{c}{Qwen3.5-4B}
& \multicolumn{2}{c}{Qwen3.5-9B}
& \multicolumn{2}{c}{R1-Qwen3-8B} \\

\cmidrule(lr){2-3}
\cmidrule(lr){4-5}
\cmidrule(lr){6-7}
\cmidrule(lr){8-9}
\cmidrule(lr){10-11}

&
Acc. & TITE
& Acc. & TITE
& Acc. & TITE
& Acc. & TITE
& Acc. & TITE \\

\midrule

Oracle
& \textbf{0.609} & --
& \textbf{0.703} & --
& \textbf{0.683} & --
& \textbf{0.743} & --
& \textbf{0.730} & -- \\

Full Thinking
& 0.559 & --
& \underline{0.611} & --
& 0.569 & --
& \underline{0.635} & --
& 0.540 & -- \\

No thinking
& 0.525 & --
& 0.536 & --
& 0.524 & --
& 0.587 & --
& 0.494 & -- \\

\midrule

Statistical EE
& 0.560 & 0.161
& 0.609 & 0.060
& \underline{0.573} & 0.003
& \underline{0.635} & \textbf{0.006}
& 0.540 & 0.045 \\

EntroCut
& 0.556 & 0.144
& 0.595 & 0.096
& 0.459 & 0.000
& 0.585 & 0.004
& 0.563 & 0.114 \\

DEER
& 0.559 & 0.157
& 0.594 & 0.095
& 0.465 & 0.000
& 0.581 & 0.002
& \underline{0.565} & \underline{0.121} \\

DTSR-1
& \underline{0.561} & \underline{0.162}
& 0.600 & 0.086
& 0.490 & 0.000
& 0.547 & 0.004
& 0.541 & 0.114 \\

CoDE-Stop
& 0.559 & 0.157
& 0.593 & 0.101
& 0.455 & 0.000
& 0.573 & \underline{0.005}
& 0.550 & \textbf{0.141} \\

\midrule

DynaSor
& 0.554 & 0.137
& 0.610 & 0.068
& 0.510 & 0.000
& 0.553 & 0.000
& 0.536 & 0.075 \\

PUMA
& 0.550 & 0.137
& 0.605 & \underline{0.110}
& 0.517 & 0.000
& 0.605 & 0.002
& 0.559 & 0.101 \\

Answer Convergence
& 0.560 & 0.155
& 0.609 & 0.084
& 0.533 & 0.001
& 0.595 & 0.002
& 0.541 & 0.054 \\

\midrule

TERMINATOR
& 0.458 & 0.066
& 0.549 & 0.036
& 0.531 & \underline{0.016}
& 0.431 & 0.000
& 0.335 & 0.010 \\

Thought calibration
& 0.560 & 0.158
& 0.602 & \textbf{0.119}
& 0.492 & \textbf{0.044}
& 0.423 & 0.000
& 0.538 & 0.047 \\

\midrule

Exit agreement
& \underline{0.561} & \textbf{0.241}
& 0.598 & 0.084
& 0.498 & 0.000
& 0.585 & \textbf{0.006}
& 0.522 & 0.079 \\

\bottomrule
\end{tabular}%
}
\end{table*}

Tables~\ref{tab:acc-tite-math-qa},
\ref{tab:acc-tite-code}, and
\ref{tab:acc-tite-hallucination} report accuracy
and TITE across five models and three task domains.
Since TITE is computed independently for each benchmark
and then macro-averaged, a method may obtain a positive
TITE even when its average accuracy falls below the baseline.

The discrete-budget oracle consistently achieves
higher accuracy than natural full reasoning.
On mathematical and scientific reasoning,
gpt-oss-20b improves from $0.824$ to $0.894$,
while Gemma4 E4B improves from $0.604$ to $0.703$.
A similar pattern appears on code generation,
where R1-Qwen3-8B improves from $0.382$ to $0.514$,
and on hallucination benchmarks, where its accuracy
increases from $0.540$ to $0.730$.

\begin{figure*}[t]
    \centering
    \includegraphics[width=\textwidth]{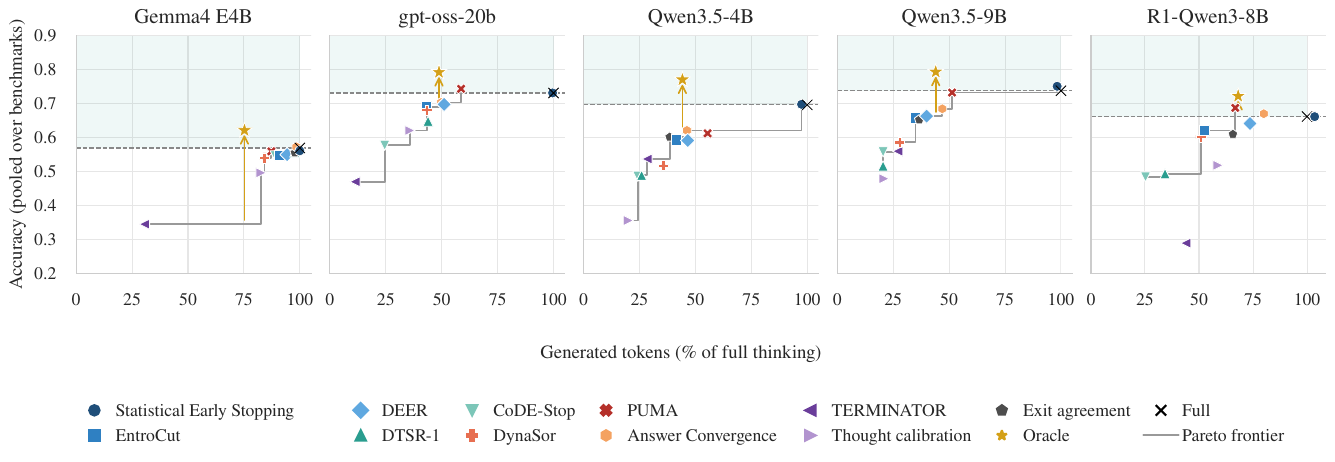}
    \caption{
    \textbf{Accuracy versus total generated tokens
    across five reasoning models.}
    The dashed line indicates full-reasoning
    accuracy, while the shaded region represents
    lower-cost operating points with at least
    the same accuracy.
    Stars indicate the discrete-budget oracle,
    and arrows illustrate the remaining oracle gap.
    }
    \label{fig:pareto}
\end{figure*}

Figure~\ref{fig:pareto} illustrates the relationship
between accuracy and total generated tokens,
normalized to full reasoning.
Across all five models and three task domains,
the oracle achieves higher accuracy at lower token
cost than full reasoning.
Statistical Early Exit generally maintains accuracy
close to full reasoning while providing little
reduction in total token cost.
In contrast, more aggressive methods such as
DTSR-1, CoDE-Stop, and some training-based approaches
achieve lower token costs but often suffer
substantial accuracy degradation.
DEER, EntroCut, and PUMA exhibit more favorable
accuracy-cost trade-offs in several settings,
although none consistently dominates across
all models and task domains.
The gap between these methods and the oracle
suggests that current approaches still struggle
to identify reliable stopping positions while
keeping overall computational costs low.

\paragraph{High accuracy does not guarantee token efficiency.}

On gpt-oss-20b mathematical and scientific reasoning,
Statistical Early Exit achieves $0.824$ accuracy
but only $0.191$ TITE.
DEER and EntroCut obtain lower accuracies of
$0.773$, yet reach considerably higher TITE
scores of $0.381$ and $0.380$, respectively.
Similarly, on Qwen3.5-4B, Statistical Early Exit
preserves full-reasoning accuracy at $0.802$
with a TITE of $0.151$, whereas EntroCut
achieves $0.269$ TITE at $0.700$ accuracy. Answer-based approaches also face this trade-off.
For example, PUMA reaches $0.799$ accuracy
on gpt-oss-20b mathematical and scientific reasoning,
exceeding DEER, but achieves a lower TITE of $0.278$.

On R1-Qwen3-8B, Thought Calibration achieves
higher accuracy than TERMINATOR across all three
domains, but its TITE remains limited:
$0.189$ on mathematical and scientific reasoning,
$0.044$ on code generation, and $0.047$ on
hallucination benchmarks.
On mathematical and scientific reasoning,
TERMINATOR achieves $0.286$ accuracy and $0.017$ TITE,
compared with $0.766$ accuracy under full reasoning.

\paragraph{Efficiency is strongly dependent on the task domain.}

The effectiveness of early stopping varies
considerably across domains.
Mathematical and scientific reasoning generally
shows the strongest efficiency, with the best
TITE scores ranging from $0.207$ to $0.406$
across the five models.
In comparison, code generation remains more
challenging. Except for Gemma4 E4B, the highest
TITE scores are below $0.15$, despite the
substantial accuracy improvements demonstrated
by the oracle.

Hallucination benchmarks exhibit similarly weak
efficiency, particularly on Qwen3.5-4B and
Qwen3.5-9B, where the best TITE scores reach
only $0.044$ and $0.006$, respectively.
Several methods even perform below the
no-thinking accuracy baseline while consuming
more tokens. For example, on Qwen3.5-4B
hallucination benchmarks, EntroCut and DEER
achieve accuracies of $0.459$ and $0.465$,
respectively, compared with $0.524$ without
reasoning, despite using substantially more tokens.

\subsection{RQ2: Correctness generally improves with reasoning, but is not strictly monotonic}
\label{subsec:rq2_preservation}

\begin{table*}[t]
\centering
\small
\caption{Continuation reasoning-error rate (CER, \%). Lower is better.}
\label{tab:continuation-error}

\begin{tabular}{lccccc}
\toprule
Benchmark
& Gemma4 E4B
& gpt-oss-20b
& Qwen3.5-4B
& Qwen3.5-9B
& R1-Qwen3-8B \\
\midrule

AIME26
& 9.63
& 0.83
& 3.97
& 3.99
& 1.40 \\

GPQA-Diamond
& 19.46
& 13.33
& 6.10
& 5.95
& 17.34 \\

GSM8K
& 6.76
& 1.36
& 5.66
& 7.80
& 1.60 \\

HaluEval
& 3.13
& 6.57
& 6.80
& 4.97
& 20.59 \\

LiveCodeBench
& 8.60
& 9.66
& 9.26
& 8.28
& 17.90 \\

MATH500
& 5.22
& 3.62
& 2.94
& 3.74
& 1.57 \\

MMLU-Pro
& 11.08
& 6.00
& 11.83
& 6.88
& 13.06 \\

TruthfulQA
& 12.48
& 10.05
& 16.87
& 12.96
& 13.11 \\

\bottomrule
\end{tabular}
\end{table*}

The continuation-error analysis shows that reaching a correct
answer does not guarantee that subsequent reasoning will
preserve it. The frequency of such errors varies considerably
across benchmarks and models.
On gpt-oss-20b, continuation-error rates remain below
$4\%$ on AIME26, GSM8K, and MATH500, but increase to
$13.33\%$ on GPQA-Diamond, $10.05\%$ on TruthfulQA,
and $9.66\%$ on LiveCodeBench.
A similar pattern appears on R1-Qwen3-8B, where
continuation-error rates remain below $2\%$ on
AIME26 and MATH500, but reach $20.59\%$ on HaluEval
and $17.90\%$ on LiveCodeBench.
However, this pattern does not hold across all models.
On Gemma4 E4B, continuation-error rates range from
$3.13\%$ on HaluEval to $19.46\%$ on GPQA-Diamond.

\paragraph{Random-exit validation.}

\begin{figure}[ht]
    \centering
    \includegraphics[width=0.85\columnwidth]{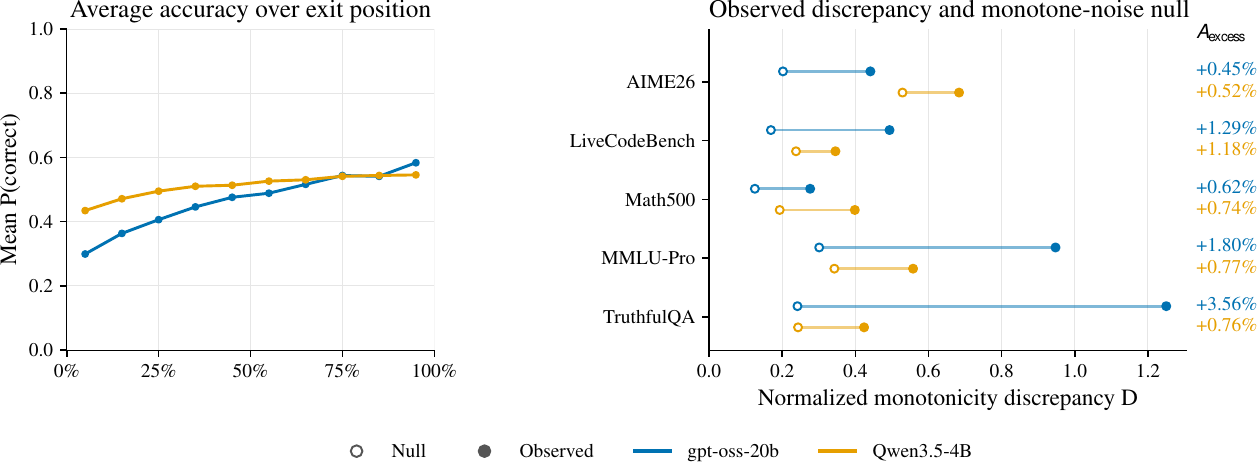}

    \caption{(a) Averaged over trajectories. (b) Per trajectory, the observed monotonicity discrepancy $D_{\mathrm{obs}}$ exceeds the value expected under a monotone model with decoding noise $D_{\mathrm{null}}$ in every setting. The right column gives the excess reversal magnitude $A_{\mathrm{excess}}$.} 
    \label{fig:avg_perf}
\end{figure}

We further evaluate randomly sampled exit positions with repeated generations as a separate validation. Although average performance increases with reasoning budget, individual trajectories exhibit more reversals than expected from decoding noise alone  (Figure \ref{fig:avg_perf}). Across the tested settings, the observed discrepancy exceeds the fitted monotone baseline, with the overall is
$D_{\mathrm{overall}}=0.5871$, and the average excess
reversal after subtracting the amount expected from decoding noise is
$1.169\%$ . The effect is especially large
for gpt-oss-20b on TruthfulQA, where excess reversal reaches $3.56\%$.

\paragraph{Exit-trap analysis.}

To understand how non-monotonic reasoning affects early stopping, we use the term \textit{exit traps} for a descriptive analysis of unfavorable stopping positions and unstable correctness patterns. We classify failed exits into three types of traps, demonstrated in Figure~\ref{fig:trace}:

\begin{itemize}
    \item \textbf{Exit too early:}
    The method terminates reasoning before a reliable
    answer has been reached.

    \item \textbf{Reasoning collapse:}
    The method continues beyond a correct exit point,
    after which the reasoning trajectory becomes
    incorrect and does not recover a correct answer
    at any later evaluated position. 

    \item \textbf{Noisy:}
    The reasoning trajectory frequently alternates
    between correct and incorrect answers. We
    classify a trajectory as noisy when it has
    at least ten correctness shifts within the window size of 32 consecutive exit position. Detailed in Appendix~\ref{sec:noisy_selection}
\end{itemize}

\begin{figure*}[!t]
    \centering
    \includegraphics[
        width=\textwidth
    ]{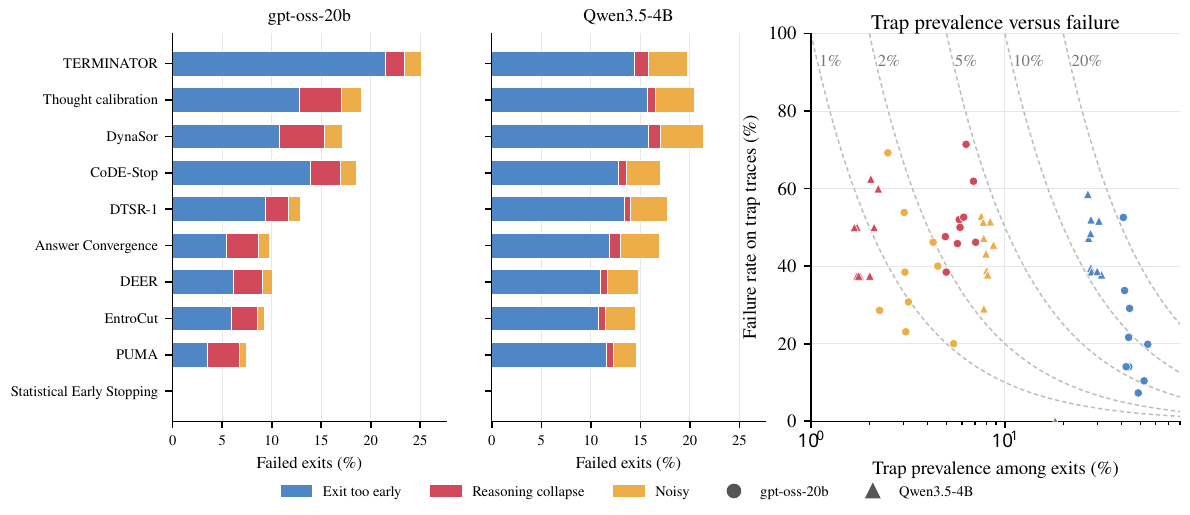}

    \caption{Exit traps. (a, b) Share of each method's exits that fail because of each trap. (c) Each point is one (method, trap) pair: trap prevalence among the method's exits versus the failure rate on trapped traces. Dashed curves are iso-lines of total failure.}
    \label{fig:classification_percentage}
\end{figure*}

Figure~\ref{fig:classification_percentage} shows that exiting too early is the dominant failure mode for most early-stopping methods on both gpt-oss-20b and Qwen3.5-4B, particularly for Terminator, Thought Calibration, DynaSor, and CoDE-Stop. In contrast, methods such as DEER,  Answer Convergence, and EntroCut exhibit smaller premature-exit shares. Reasoning collapse and noisy trajectories contribute less to the overall distribution but remain challenging for stopping methods. Figure~\ref{fig:classification_percentage}(c) further shows that trap prevalence and conditional failure rate capture different aspects of stopping difficulty. While early exits account for a larger share of failures, reasoning collapse and noisy trajectories can exhibit high conditional failure rates despite their lower prevalence. The dashed iso-failure curves illustrate that a trap's overall contribution depends on both its prevalence and the probability of failure when encountered.

\paragraph{Qualitative trace analysis.}

To investigate the reasoning behaviors behind these traps, we conduct a small qualitative analysis of 40 sampled trajectories, human–AI collaborative annotation process with GPT-6 Astra \cite{openai2026astra} to identify reasoning advancement, verification, reconsideration, and errors between exit checkpoints. In early-exit cases, most traces first express the correct result or an equivalent diagnosis near the examined boundary, suggesting that premature stopping often interrupts reasoning before a useful result is established. In contrast, noisy trajectories do not necessarily involve changes in the model's written conclusion: about 40\% of the examined correctness reversals retain the same candidate answer. For collapse, 4/14 traces visibly abandon an earlier candidate, while 7 retain their previous selection and 3 remain ambiguous. Detail can be found in Appendix~\ref{sec:noisy_explain}.

Figure~\ref{fig:random_text} illustrates a case of harmful reconsideration. For TruthfulQA question 229, the model initially identifies both A and B as correct. However, further reasoning introduces a possible trade-secret exception, leading it to question A and eventually select only B. Consequently, forced-exit accuracy drops and does not recover at later evaluated checkpoints. 

\subsection{RQ3: Reliability Beyond Correctness}
\label{subsec:rq4_reliability}

\subsubsection{Post-Stop Reasoning Continuation}
\label{subsubsec:post_stop_results}

\begin{figure*}[ht]
    \centering
    \includegraphics[width=\textwidth]{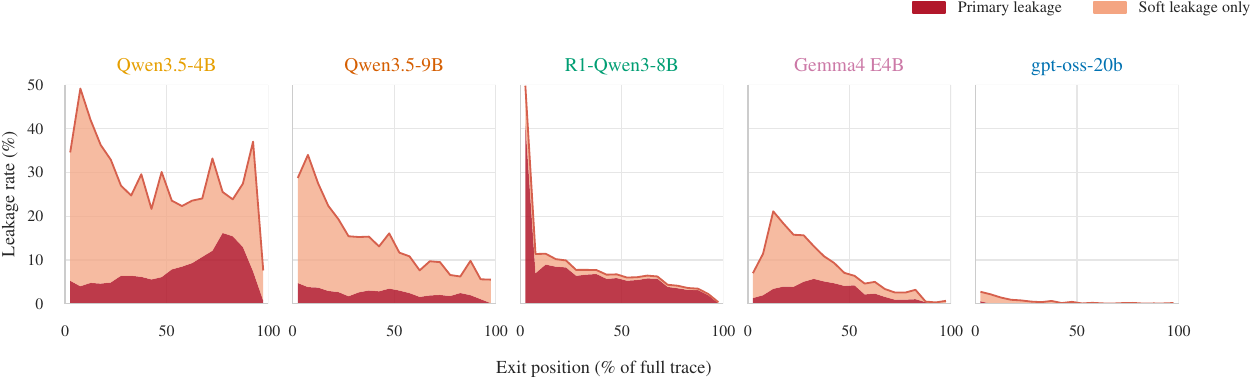}
    \caption{Post-stop reasoning leakage rate by normalized exit position, split into primary leakage and soft leakage. Bin width is 0.05.}
    \label{fig:leakage}
\end{figure*}

We measure post-stop reasoning leakage by identifying two types of
continuation: \emph{primary leakage}, where explicit thinking tags
reappear after the forced stop, and \emph{soft leakage}, where
continuation markers suggest that reasoning persists. Figure~\ref{fig:leakage} reports their
rates across normalized exit positions.

\paragraph{Leakage varies substantially across models.}
Qwen3.5-4B exhibits the highest and most persistent leakage with nearly $50\%$ at early and remaining substantial
across much of the reasoning trajectory. Qwen3.5-9B also shows
considerable leakage at early exits, but its rate generally
declines as the exit position increases. R1-Qwen3-8B displays
a sharp initial peak, followed by a rapid reduction in leakage.
Gemma4 E4B shows a smaller early peak of approximately $20\%$,
with leakage approaching zero at later positions. In contrast,
gpt-oss-20b exhibits almost no post-stop continuation throughout
the trajectory.

Across most models and exit positions, soft-only leakage accounts for a large portion of observed continuation, primary leakage nevertheless remains visible, especially for Qwen3.5-4B and R1-Qwen3-8B. Although leakage generally decreases with exit position for several models, Qwen3.5-4B exhibits a renewed increase toward later exits. Thus, allowing more reasoning before stopping does not necessarily eliminate post-stop continuation.

\subsubsection{Retained Justification Quality}
\label{subsubsec:justification_results}

\begin{table*}[t]
\centering
\caption{Justification quality by outcome correctness, relative to
\textsc{Full Thinking} on the same model and benchmark. Full Thinking
rows report the mean judge score ($10$--$100$); other rows report the
difference from Full Thinking.}
\label{tab:justification_quality}
\small
\setlength{\tabcolsep}{4pt}

\footnotesize
\setlength{\tabcolsep}{2.5pt}
\renewcommand{\arraystretch}{0.95}

\begin{tabular}{llrrrrrrrr}
\toprule
\textbf{Model} & \textbf{Method}
& \multicolumn{2}{c}{\textbf{GPQA-Diamond}}
& \multicolumn{2}{c}{\textbf{LiveCodeBench}}
& \multicolumn{2}{c}{\textbf{MMLU-Pro}}
& \multicolumn{2}{c}{\textbf{TruthfulQA}} \\
\cmidrule(lr){3-4}
\cmidrule(lr){5-6}
\cmidrule(lr){7-8}
\cmidrule(lr){9-10}
& & $J_{\mathrm{correct}}$ & $J_{\mathrm{incorrect}}$
  & $J_{\mathrm{correct}}$ & $J_{\mathrm{incorrect}}$
  & $J_{\mathrm{correct}}$ & $J_{\mathrm{incorrect}}$
  & $J_{\mathrm{correct}}$ & $J_{\mathrm{incorrect}}$ \\
\midrule

\multirow{5}{*}{gpt-oss-20b}
& Full Thinking & 85.2 & 43.6 & 88.4 & 77.2 & 89.4 & 53.0 & 88.8 & 73.4 \\
& EntroCut   & \negd{-16.3} & \negd{-20.0} & \negd{-8.2} & \negd{-12.1} & \negd{-12.6} & \negd{-16.2} & \negd{-15.3} & \negd{-23.6} \\
& DTSR-1       & \negd{-15.0} & \negd{-20.4} & \negd{-9.9} & \negd{-13.4} & \negd{-11.2} & \negd{-21.3} & \negd{-10.0} & \negd{-18.2} \\
& DEER       & \negd{-18.9} & \negd{-19.7} & \negd{-5.6} & \negd{-8.8}  & \negd{-10.3} & \negd{-19.2} & \negd{-9.8}  & \negd{-26.4} \\
& TERMINATOR & \negd{-59.9} & \negd{-29.2} & \negd{-8.3} & \negd{-13.6} & \negd{-41.4} & \negd{-38.0} & \negd{-60.1} & \negd{-50.7} \\
\midrule

\multirow{5}{*}{Qwen3.5-4B}
& Full Thinking & 93.9 & 39.4 & 90.2 & 52.5 & 95.8 & 74.4 & 90.8 & 61.1 \\
& EntroCut   & \negd{-5.1}  & \negd{-3.8}  & \negd{-9.5}  & \negd{-17.8} & \negd{-8.4}  & \negd{-16.0} & \negd{-14.0} & \negd{-14.7} \\
& DTSR-1       & \negd{-6.3}  & \negd{-1.8}  & \negd{-14.9} & \negd{-11.7} & \negd{-9.6}  & \negd{-28.3} & \negd{-20.0} & \negd{-23.4} \\
& DEER       & \negd{-3.6}  & \negd{-3.2}  & \negd{-6.5}  & \negd{-13.6} & \negd{-10.7} & \negd{-13.8} & \negd{-10.3} & \negd{-12.5} \\
& TERMINATOR & \negd{-10.2} & \negd{-9.3}  & \negd{-25.9} & \negd{-13.9} & \negd{-11.1} & \negd{-31.9} & \negd{-73.0} & \negd{-43.6} \\
\midrule

\multirow{5}{*}{Qwen3.5-9B}
& Full Thinking & 92.2 & 47.3 & 93.8 & 70.5 & 93.1 & 77.7 & 93.7 & 80.5 \\
& EntroCut   & \negd{-4.9}  & \negd{-7.5}  & \negd{-12.7} & \negd{-27.8} & \negd{-7.5} & \negd{-16.7} & \negd{-26.8} & \negd{-33.5} \\
& DTSR-1       & \negd{-1.6}  & \negd{-14.7} & \negd{-17.1} & \negd{-24.0} & \negd{-5.7} & \negd{-38.0} & \negd{-31.8} & \negd{-43.7} \\
& DEER       & \negd{-4.5}  & \negd{-6.7}  & \negd{-7.2}  & \negd{-17.8} & \negd{-7.5} & \negd{-26.7} & \negd{-27.0} & \negd{-37.3} \\
& TERMINATOR & \posd{2.8}   & \negd{-1.1}  & \negd{-28.4} & \negd{-23.1} & \negd{-8.4} & \negd{-32.1} & \negd{-27.6} & \negd{-38.1} \\
\midrule

\multirow{5}{*}{R1-Qwen3-8B}
& Full Thinking & 76.2 & 44.1 & 66.8 & 45.5 & 89.1 & 59.2 & 87.2 & 73.3 \\
& EntroCut   & \negd{-8.1}  & \negd{-16.2} & \negd{-4.8}  & \negd{-11.5} & \negd{-9.5} & \negd{-19.5} & \negd{-24.5} & \negd{-30.9} \\
& DTSR-1       & \negd{-20.1} & \negd{-20.4} & \negd{-9.7}  & \negd{-15.1} & \negd{-5.2} & \negd{-17.3} & \negd{-25.5} & \negd{-34.8} \\
& DEER       & \negd{-5.6}  & \negd{-14.7} & \negd{-3.9}  & \posd{2.8}   & \negd{-5.9} & \negd{-19.0} & \negd{-23.5} & \negd{-30.2} \\
& TERMINATOR & \negd{-13.7} & \negd{-9.9} & \negd{-28.6} & \negd{-21.3} & \negd{-2.1} & \negd{-7.4} & \negd{-15.0} & \negd{-35.8} \\
\bottomrule
\end{tabular}
\end{table*}

\begin{table}[t]
\centering
\small
\setlength{\tabcolsep}{2pt}
\caption{Failure-mode distribution among low-quality outputs ($J<50$),
pooled across all models and benchmarks. Percentages are computed within each method's low-quality outputs.}
\label{tab:justification_failure_modes}
\begin{tabular}{lrrrrrr}
\toprule
\textbf{Method} & \textbf{Insuff.} & \textbf{Unsupp.} & \textbf{Circ.} & \textbf{Contra.} & \textbf{Other} \\
\midrule
Full Thinking  & 36.1 & 9.3 & 2.9 & 49.0 & 2.3 \\
DEER           & 38.8 & 3.3 & 1.4 & 54.4 & 1.8 \\
EntroCut       & 44.7 & 4.1 & 1.4 & 48.0 & 1.3 \\
DTSR-1           & 52.2 & 1.5 & 0.9 & 44.6 & 0.7 \\
TERMINATOR     & 71.7 & 1.1 & 0.5 & 26.1 & 0.5 \\
\bottomrule
\end{tabular}
\end{table}

Most early-stopping methods show a negative justification
delta relative to Full Thinking, and incorrect answers frequently suffer larger justification losses. DEER tends to have
the smallest loss of the three training-free methods, while EntroCut
and DTSR-1 trail it. The gap widens on TruthfulQA, where every method on
every model loses at least $9.8$ points.

TERMINATOR is the most variable method. It collapses on gpt-oss-20b
(GPQA-Diamond $-59.9$, MMLU-Pro $-41.4$, TruthfulQA $-60.1$) and on
Qwen3.5-4B TruthfulQA ($-73.0$), far beyond the loss of any other
method on the same benchmark. On Qwen3.5-9B and
R1-Qwen3-8B, its correct-output losses stay between
$+2.8$ and $-28.6$, although its incorrect-output
losses still reach $-38.1$ and $-35.8$, respectively.

Excluding TERMINATOR, correct-output losses are mostly $4$--$17$
points on LiveCodeBench and $5$--$13$ on MMLU-Pro, but reach $10$--$32$
on TruthfulQA. GPQA-Diamond shows the strongest model dependence:
losses are at most $6.3$ on Qwen3.5, $15$--$19$ on gpt-oss-20b, and
$5.6$--$20.1$ on R1-Qwen3-8B. TruthfulQA losses also differ across
models for the same method (DTSR-1: $-20.0$ on Qwen3.5-4B, $-31.8$ on
Qwen3.5-9B, $-25.5$ on R1-Qwen3-8B), so the loss depends jointly on
method, benchmark, and model.

\paragraph{Failure modes and the base-model floor.}
\textit{Insufficient derivation} and \textit{contradicts final answer}
dominate low-quality outputs for every method; no other category
exceeds $9.3\%$. TERMINATOR skews furthest toward insufficient
derivation ($71.7\%$), consistent with its collapse cells, while DEER
tracks Full Thinking most closely ($38.8\%$ vs.\ $36.1\%$ insufficient
derivation). The same two failure types also dominate Full Thinking's low-quality outputs (36.1\% insufficient derivation, 49.0\% contradiction), even though it is never early-stopped. Weak justification is therefore not caused by early stopping alone. Better thresholds can reduce errors from premature termination, but cannot fix derivation gaps already present in full reasoning.

\section{Findings}
\label{sec:key_findings}

\paragraph{The potential of early stopping remains largely unrealized.}
Across five models and three task domains, the discrete-budget
hindsight oracle achieves higher accuracy with fewer tokens than
full reasoning. Yet, after accounting for verification and answer
generation costs, the highest TITE score achieved by any evaluated
policy is only $0.468$. Existing methods tend to preserve accuracy
at the expense of efficiency or reduce computation at the expense
of accuracy, with no method consistently dominating across settings.

\paragraph{Effective stopping signals are highly domain-dependent.}
While mathematical and scientific reasoning achieves relatively
favorable efficiency, the best TITE scores on code generation
remain below $0.15$ for four of the five models. On factuality
benchmarks, early stopping can be worse than direct answering:
DEER and EntroCut reduce Qwen3.5-4B's accuracy from $0.524$ to
$0.465$ and $0.459$, respectively, while consuming more tokens.

\paragraph{Answer readiness is not a persistent state.}
Additional reasoning can invalidate an earlier correct answer,
with continuation-error rates reaching $20.59\%$. Randomized exit
experiments reveal an average reversal magnitude of $1.169$
percentage points beyond the fitted decoding-noise baseline.
Moreover, correctness can fluctuate even when the written
candidate answer remains unchanged, limiting the reliability
of answer convergence as a stopping criterion.

\paragraph{Answer correctness alone overestimates stopping reliability.}
Forcing an end-of-thinking token does not necessarily terminate
reasoning: continuation markers appear in nearly $50\%$ of early
exits for Qwen3.5-4B. Even among correct answers, retained reasoning
receives lower justification scores than full reasoning in 63 of
64 evaluated comparisons. Early stopping can therefore preserve
measured accuracy while compromising the completeness of the
reasoning that supports it.
\section{Conclusion}

We benchmark ten early-stopping methods across five reasoning models and eight tasks using common replay evaluation and the oracle-relative TITE metric. Many evaluated policies achieve low TITE scores relative to the discrete-budget oracle, especially on code and factuality/hallucination benchmarks. Randomized exit tests reveal non-monotone correctness beyond decoding noise; reliability audits find post-stop reasoning and weaker retained justification. Future stopping rules should account for correctness stability and verifier cost alongside termination behavior.

\section*{Limitations}

Our main limitation is the computational cost of evaluating early-stoppings in detail. Each additional exit position requires a new answer continuation, and repeated sampling, requiring up to 512 answer continuations for a single trajectory. Within the available compute budget, the reported monotonicity validation covers only sub-sampled data and models.

Our cost measurements  assume ideal KV-cache reuse and exclude the overhead of cache management, therefore do not directly establish reductions in latency, memory use, or energy consumption, as different architectures, quantization methods produce different results and favor different types of hardware. Those measurements would require deployment experiments that account for hardware, batching, and scheduling.

Finally, the reliability measures also have limits: lexical indicators can misidentify post-stop reasoning, and automated justification scores assess the visible trace without establishing whether it faithfully represents the model's internal computation.

\section*{Acknowledgments}

The authors gratefully acknowledge VinUniversity
for its institutional support for this research.


\bibliographystyle{ACM-Reference-Format}
\bibliography{custom.bib}

\appendix

\section{Early-stopping Method Configurations}
\label{app:method-configurations}

\subsection{Exit Agreement Configurations}
\label{app:exit-agreement}

The stage boundaries are $1{,}024$, $2{,}048$, and $4{,}096$
reasoning tokens. Votes accumulate across stages. Once the
quorum is reached, the policy returns the latest eligible
answer, or waits for the first eligible answer within the
current stage. Voting ends strictly before $8{,}192$ tokens
or the recorded hard stop, whichever occurs first. If no
exit is selected, the recorded 16K-budget answer is returned.

We evaluate $\rho\in\{0.50,0.75,1.00\}$ over five voter sets:
\begin{itemize}
    \item CoDE-Stop, DTSR-1, DEER, EntroCut;
    \item DTSR-1, DEER, EntroCut;
    \item CoDE-Stop, DEER, EntroCut;
    \item CoDE-Stop, DTSR-1, DEER, EntroCut, TERMINATOR;
    \item DEER, EntroCut.
\end{itemize}

Among configurations with the widest benchmark coverage,
selection maximizes benchmark-average TITE, with ties broken
by lower mean token cost. DEER and EntroCut are selected
for all models, with $\rho=0.75$ for Gemma4 E4B and
Qwen3.5-9B, and $\rho=0.50$ for the remaining models.

\subsection{Individual Method Configurations}
\label{app:individual-replay-methods}

DEER, EntroCut, CoDE-Stop, and DTSR-1 use the reflection
markers \texttt{wait}, \texttt{Wait}, \texttt{but}, and
\texttt{But}. The remaining method-specific settings are
as follows:

\begin{description}
    \item[DEER.] Confidence threshold $0.95$, using the
    geometric mean of maximum token probabilities.

    \item[EntroCut.] Mean entropy threshold $0.20$ nats
    and residual probability mass threshold $0.05$.

    \item[CoDE-Stop.] Confidence threshold $0.95$ or
    degeneration threshold $1.0$, with five log probabilities
    per token. 

    \item[DTSR-1.] Sufficiency threshold $80$ and minimum
    checkpoint interval of 128 tokens.

    \item[PUMA.] Semantic steps of 200 characters,
    redundancy threshold $0.35$, confidence threshold
    $0.90$, and $\epsilon=0.05$. Two consecutive
    confirmations, window size one, redundancy limit 40,
    and 15-token probes with three log probabilities.

    \item[Statistical Early Exit.] Uncertainty density
    threshold $0.025$, evaluated every 256 tokens.

    \item[DynaSor.] Checkpoint interval of 128 tokens,
    five-answer window, and consistency threshold $0.80$.

    \item[Answer Convergence.] Checkpoint interval of
    128 tokens, requiring ten consecutive matching answers.

    \item[TERMINATOR.] Classification window of 10 tokens,
    checkpoint interval of 16 tokens, window threshold
    $0.51$, and per-token threshold $0.70$.

    \item[Thought Calibration.] Checkpoint interval of
    64 tokens and calibrated risk tolerance $0.05$.
\end{description}

\section{Trap in Reasoning Trace}

\subsection{Threshold selection for Noisy Trap}
\label{sec:noisy_selection}

\begin{figure*}[!t]
    \centering
    \includegraphics[
        width=0.4\textwidth
    ]{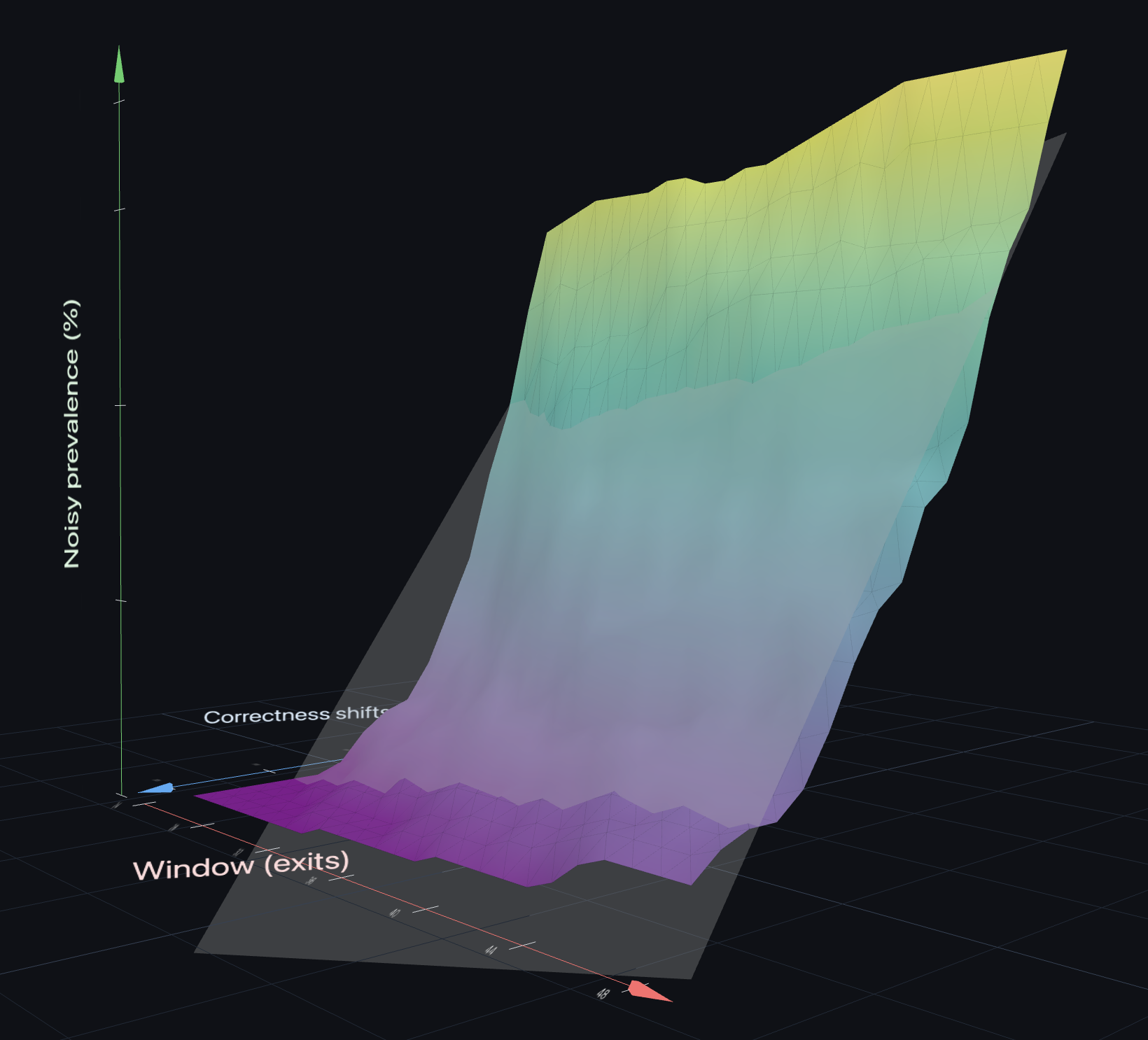}
    \includegraphics[
        width=0.5\textwidth
    ]{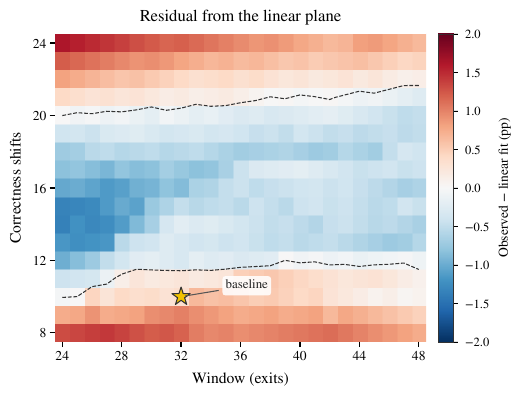}

    \caption{Sensitivity of Noisy prevalence to the window size $w$ and the minimum number of signal changes $k$. The left panel shows the observed prevalence surface and its linear fit, while the right panel shows the residuals from the linear model.}
    \label{fig:threshold_selection}
\end{figure*}

We define noisy trap as the occurrence of at least $k$ signal changes within a window of $w$ exits. To examine the sensitivity of this definition to the choice of thresholds, we conduct a grid search over $k \in [8,24]$ and $w \in [24,48]$.

As shown in Fig.~\ref{fig:threshold_selection}, the number of noisy samples increases with larger window sizes $w$ and smaller thresholds $k$. This relationship is approximately achieving a $R^2 = 0.9084$ with linear regression and $R^2 = 0.9904$ in a quadratic fit, but the fitted surface does not exhibit an interior maximum. Prevalence changes smoothly with the thresholds, without a clear optimal combination or distinct transition point. We therefore adopt $k = 10$ and $w = 32$ as practical reference thresholds for the analysis.

\subsection{Lexical Pattern in each trap}
\begin{figure}[htbp]
    \centering

        \includegraphics[width=0.6\linewidth]{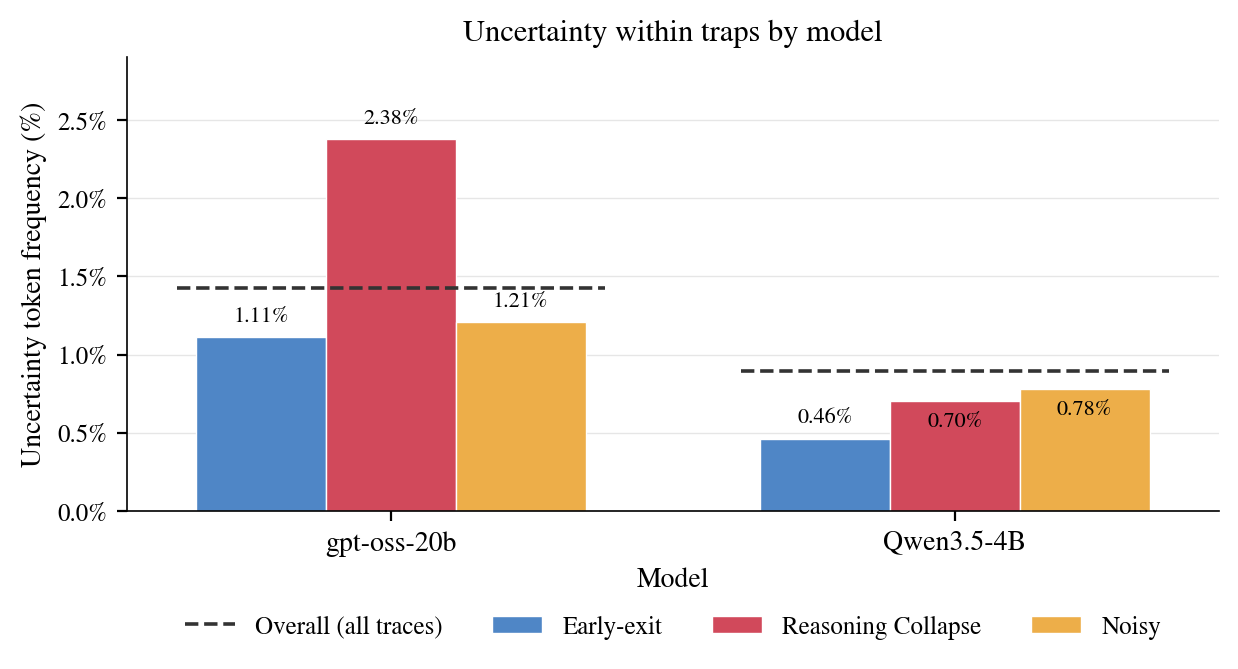}
        \caption{Percentage of uncertainty token for each trap}
        \label{fig:small}

\end{figure}

In this section, we try to find any pattern to distinguish each trap  Figure~\ref{fig:small} shows that early-exit traces have the lowest
marker frequency for both models (1.11\% for gpt-oss-20b and 0.46\%
for Qwen3.5-4B). The highest frequency occurs in collapse traces for
gpt-oss-20b (2.38\%) and in noisy traces for Qwen3.5-4B (0.78\%).
The difference in ordering  suggests that the trap characteristics are model-dependent, however, this should not be a reliable method for classification.

\subsection{Qualitative Analysis of Exit Traps}
\label{sec:noisy_explain}

\paragraph{Objective and methodology.}
To better understand the reasoning behaviors associated with
the three exit traps, we conduct an exploratory
qualitative analysis of 40 reasoning trajectories
(12 early-exit, 14 noisy, and 14 collapse).
We use a human--AI collaborative annotation process
assisted by GPT-6 Astra~\cite{openai2026astra}
to examine reasoning changes between exit checkpoints by first generating coding draft by AI then verified by human.
Each segment is classified by its primary reasoning action:
\textit{advance} (new useful reasoning),
\textit{check} (valid verification),
\textit{reconsider} (exploring alternatives),
\textit{continue} (repeating existing reasoning),
or \textit{error} (invalid inference).
Ambiguous segments are marked as unclear.
Reviewers annotate reasoning text with trap labels and
correctness scores masked. 

\paragraph{Findings.}
Table~\ref{tab:qualitative_traps} summarizes the
observed reasoning behaviors across the three traps.

\begin{table*}[t]
\centering
\small
\caption{Qualitative comparison of reasoning behaviors
across exit traps. Counts refer to the reviewed sample.}
\label{tab:qualitative_traps}
\begin{tabular}{p{0.16\textwidth}
                p{0.38\textwidth}
                p{0.36\textwidth}}
\toprule
\textbf{Trap} &
\textbf{Observed behavior} &
\textbf{Supporting evidence} \\
\midrule

\textbf{Early exit}
&
Reasoning often stops before a useful result
or necessary reasoning step is completed.
Additional thinking allows the model to
develop or verify its solution.
&
8/12 first express the correct result or
equivalent diagnosis near the examined boundary.
Reasoning advancement is the most common action
(7/12).
\\

\addlinespace

\textbf{Noisy}
&
Correctness fluctuates without necessarily
changing the written answer.
Additional reasoning frequently checks or
reconsiders existing candidates.
&
23/56 examined reversals retain the same
written candidate. Only two explicit answer
switches are observed, both in one trajectory.
\\

\addlinespace

\textbf{Collapse}
&
Some trajectories abandon an earlier correct
candidate after reconsideration, while others
retain the same written answer despite
declining forced-exit accuracy.
&
4/14 visibly move away from an earlier
candidate, 7/14 retain their previous
selection, and 3/14 remain ambiguous.
\\

\bottomrule
\end{tabular}
\end{table*}

Figure~\ref{fig:trace} illustrates a collapse case
where the model initially considers both A and B
correct, but later introduces a trade-secret
exception and rejects A, eventually selecting B.
This demonstrates how further reconsideration
can undermine an earlier correct solution.

However, such revisions do not explain every
correctness reversal. Across 38 matched
within-rollout comparisons, checking or continuing
an existing solution occurs in 25/38 stable windows,
compared with 14/38 trap windows.
Moreover, noisy and collapse cases frequently
retain the same written answer despite changes
in forced-exit correctness. These  suggest that early exits are often
associated with incomplete reasoning, while
noisy and collapse trajectories involve a mixture
of reconsideration and forced-answer instability.

\end{document}